%% file: ISM-CVPR-arXiv.tex
\documentclass[12pt]{article}

\input{math_commands.tex}

\usepackage{graphicx}
\usepackage{amsmath}
\usepackage{amssymb}
\usepackage{amsthm}
\usepackage{booktabs}
\usepackage{cmll}
\usepackage{mathtools}
\usepackage{subcaption}
\usepackage{algorithm,algorithmicx,algpseudocode}
\usepackage{xr}
\usepackage{authblk}
\usepackage{geometry}
\usepackage[pagebackref,breaklinks,colorlinks]{hyperref}

\usepackage[capitalize]{cleveref}
\crefname{section}{Sec.}{Secs.}
\Crefname{section}{Section}{Sections}
\Crefname{table}{Table}{Tables}
\crefname{table}{Tab.}{Tabs.}

\newtheorem{theorem}{\bf{Theorem}}

\newtheorem{lemma}{\bf{Lemma}}
\newtheorem{definition}{\bf{Definition}}

\newtheorem{example}{\bf{Example}}

\def\cvprPaperID{11673} 
\def\confName{CVPR}
\def\confYear{2022}
\makeatletter
\def\ps@IEEEtitlepagestyle{%
	\def\@oddfoot{\mycopyrightnotice}%
	\def\@evenfoot{}%
}
\def\mycopyrightnotice{%
	{\hfill \footnotesize 978-1-4673-9563-2/15/\$31.00 \copyright 2015 IEEE\hfill}
}
\makeatother
\usepackage{fancyhdr}
\renewcommand{\headrulewidth}{0pt}
\renewcommand{\footrulewidth}{0pt}
\begin{document}

\title{Out-of-distribution Generalization with Causal Invariant Transformations}

	\author[1,2]{Ruoyu Wang\thanks{Equal contribution.}}
	\author[1,2]{Mingyang Yi$^{*}$}
	\author[3]{Zhitang Chen}
	\author[3]{Shengyu Zhu\thanks{Corresponding author.}}
	\affil[1]{University of Chinese Academy of Sciences}
	\affil[2]{Academy of Mathematics and Systems Science, Chinese Academy of Sciences}
	\affil[3]{Huawei Noah's Ark Lab}

\maketitle
\thispagestyle{fancy}
\fancyhead{}
\lhead{}
\lfoot{\copyright~2022 IEEE}
\cfoot{}
\rfoot{}
\begin{abstract}
   In real-world applications, it is important and desirable to learn a model that performs well on out-of-distribution (OOD) data. Recently, causality has become a powerful tool to tackle the OOD generalization problem, with the  idea resting on the causal mechanism that is invariant across  domains of interest. To leverage the generally unknown causal mechanism, existing works assume a linear form of causal feature or require sufficiently many and diverse training domains, which are usually restrictive in practice. In this work, we obviate these assumptions and tackle the OOD problem without explicitly recovering the causal feature. Our approach is based on transformations that modify the non-causal feature but leave the causal part unchanged, which can be either obtained from prior knowledge or learned from the training data in the multi-domain scenario. Under the setting of invariant causal mechanism, we theoretically show that if all such transformations are available, then we can learn a minimax optimal model across the domains using only single domain data. Noticing that knowing a complete set of these causal invariant transformations may be impractical, we further show that it suffices to know only a subset of these transformations. Based on the theoretical findings,  a regularized training procedure is proposed to improve the OOD generalization capability. Extensive experimental results on both synthetic and real datasets verify the effectiveness of the proposed algorithm, even with only a few causal invariant transformations.
\end{abstract}

\section{Introduction}
The success of many machine learning algorithms with empirical risk minimization (ERM) relies on the independent and identically distributed (i.i.d.) hypothesis that training and test data originate from a common distribution.
In practice, however, data in different domains or environments are often heterogeneous, due to changing circumstances, selection bias, and time-shifts in the distributions \cite{meinshausen2015maximin, rothenhausler2021anchor}. Accessing data from all the domains of interest, on the other hand, is expensive or even impossible. Consequently, the problem of learning a model that generalizes well on the unseen target distributions is a practically important but also challenging task and has gained much research attention in the past decades \cite{blanchard2011generalizing,blanchard2021domain,fang2013unbiased,muandet2013domain,volpi2018generalizing}.
\par
Since data from some domains are unavailable, assumptions or prior knowledge on the unseen domains are generally required to achieve a guaranteed out-of-distribution (OOD) generalization performance. Recently, causality has become a powerful tool to tackle the OOD problem \cite{peters2017elements, rojas2018invariant,arjovsky2019invariant,rothenhausler2021anchor}. This is based on the assumption or observation that the underlying causal mechanism is invariant in general, even though the data distributions may vary with domains. It has been shown that a model would perform well across different domains in the minimax sense if such a causal mechanism is indeed captured.


\par
To capture the invariant causal mechanism,  existing works have assumed a particular form of the causal diagram \cite{rothenhausler2021anchor,subbaswamy2019preventing,mitrovic2020representation,heinze2021conditional,mahajan2021CausalMatching}, which may be restrictive in practice and is untestable from the observed data. Other works try to recover the so-called ``causal feature'' from the data to improve the OOD generalization performance \cite{rojas2018invariant,chang2020invariant,liu2021heterogeneous,gimenez2021identifying}. 
These works usually assume a linear form of causal feature \cite{chang2020invariant, rothenhausler2021anchor,liu2021heterogeneous,gimenez2021identifying} or that there are sufficiently many and diverse training domains so that the causal feature could be identified via certain invariant properties \cite{peters2016causal,rojas2018invariant,arjovsky2019invariant}. In the absence of these assumptions, existing methods such as invariant risk minimization \cite{arjovsky2019invariant} can fail to capture the invariance or recover the causal feature
even in simple examples \cite{kamath2021doesIRM}. In real applications  like image classification, the linearity assumption may not hold, and it may be expensive or even impossible to ensure that the available domains are indeed sufficient. As such, the identifiability issue of causal feature can hardly be resolved in practice. 
\par 
In this paper, we obviate the aforementioned assumptions and propose a new approach to learn a robust model for OOD generalization under the invariant causal mechanism assumption. We do not try to \emph{explicitly} recover the causal feature; rather, we directly learn a model that takes advantage of the invariant properties. Our approach is based on the observation that though the explicit functional form of the causal feature is generally unknown and maybe also hard to learn, we often have some prior knowledge on the transformations that the causal feature is invariant to, i.e., transformations that modify the input data but do not change their causal feature. For example, the shape of  digit in an image from  \texttt{MNIST} dataset \cite{lecun1998gradient} can be treated as a causal feature when predicting the digit, while flipping or rotation does not change causal meanings. A  detailed discussion on this issue is given in \Cref{subsec: prior knowledge}. We refer to these transformations as \emph{causal invariant transformations} (CITs). 
\par
Theoretically, we prove that given complete prior knowledge of CITs, it is feasible to learn a model with OOD generalization capability using only single domain data. Specifically, we show that if all the CITs are known, then minimizing the loss over all the causally invariant transformed data, which are obtained by applying the CITs to data from the given single domain, would lead to the desired model that achieves a minimax optimality across all the domains of interest. Noticing that obtaining all CITs may be impractical, we further show that, for the purpose of OOD generalization, it suffices to know only an appropriate subset of CITs, referred to as \emph{causal essential set} and is formally defined in \cref{def:causal essential set}. The learned model is then shown to generalize to  different domains if it is invariant to transformations in a causal essential set. This is different from  existing works \cite{sokolic2017generalization,sannai2019improved} that demonstrate an improved i.i.d.~generalization capability from  invariance properties. 

Following these theoretical results, we propose to regularize training with the discrepancy between the model outputs of the original data and their transformed versions from the CITs in the causal essential set, to enhance OOD generalization. The CITs can be viewed as data augmentation operations;  in this sense, our theoretical results reveal the rationale behind data augmentation in OOD problems.	Experiments on both synthetic and real-world benchmark datasets, including \texttt{PACS} \cite{li2017deeper} and \texttt{VLCS} \cite{fang2013unbiased}, verify our theoretical findings and demonstrate the effectiveness of the proposed algorithm in terms of OOD performance. {Noticeably, in some experiments, we use CycleGAN to learn the transformations between different environments, which are then used as our CITs. This is in contrast with  \cite{zhou2020learning} which conjectured that source-to-source transformation could provide little help to domain generation tasks in their approach.} 

\section{Related Work}

As data from some unseen domains are completely unavailable, assumptions or prior knowledge on the data distributions are required to guarantee a good OOD generalization performance. We will briefly review existing domain generalization methods according to these assumptions.

\paragraph{Marginal Transfer Learning}  A branch of works assume that  distributions under different domains are i.i.d.~realizations from a superpopulation of distributions and augment the original feature space with the covariate distribution \cite{blanchard2011generalizing,blanchard2021domain}. This i.i.d.~assumption on the data distribution is akin to the random effect model \cite{laird1982random,bondell2010joint} or Bayesian approach \cite{Deely1981EB,ray2020semiparametric}, but may be inappropriate when the difference between domains is irregular, e.g., different styles and backgrounds in the \texttt{PACS} and \texttt{VLCS} datasets, respectively.

\paragraph{Robust Optimization}  Existing works also consider OOD data that lie close to the training distribution in terms of probability distance or divergence, e.g., Wasserstein distance \cite{sinha2018certifying,volpi2018generalizing,lee2017minimax,yi2021improved} or $f$-divergence \cite{hu2018does,gao2020wasserstein,duchi2021learning}. They proposed to train the model via distributional robust optimization so that the model generalizes well over a set of distributions, the so-called uncertainty set \cite{ben2013robust,shapiro2017distributionally}. However, picking a suitable probability distance and range of uncertainty set in real scenarios is difficult in practice \cite{duchi2021learning}. Besides, the distributions in the uncertainty set are actually the distributions of corrupted OOD data such as adversarial sample and noisy corrupted data \cite{yi2021improved}, while the commonly encountered style-transformed OOD data are not included \cite{hendrycks2020many}.

\paragraph{Invariant Feature} 
Another branch of methods aim to seek a model with features whose (conditional) distributions are invariant across different domains. To this end, it was proposed to learn the feature representation by minimizing some loss functions involving domain scatter \cite{muandet2013domain,ghifary2016scatter,li2018domain}. Here domain scatter is a quantity characterizing the dissimilarity between (conditional) distributions in different domains, as defined in \cite{ghifary2016scatter}. \cite{li2018adversarial} and \cite{li2018deep} considered to regularize training to reduce  the maximum mean discrepancy of the feature distributions of different domains and the Jensen-Shannon divergence of the feature distributions conditional on the outcome, respectively.  The rationale behind these methods is to minimize a term that appears in the upper bound of the prediction error in unseen target domains \cite{ben2007analysis,ben2010theory, ghifary2016scatter}. Theoretically, the success of these methods hinges on the assumption that other terms in the upper bound are small enough \cite{ghifary2016scatter}. However, the implication of this assumption is usually unclear and provides little guidance for the practitioner \cite{chen2020domain}.
Although often not stated explicitly, the validity of these methods relies on the \emph{covariate shift} or \emph{label shift} assumption that  is implausible if spurious correlations occur under certain domains \cite{chen2020domain,zhou2021domain,kuang2018stable,liu2021heterogeneous}.

\paragraph{Invariant Causal Mechanism}
As in this paper, many existing works also resort to causality to study the OOD generalization problem \cite{rojas2018invariant,arjovsky2019invariant,chang2020invariant,mitrovic2020representation,ahuja2020invariant,liu2021heterogeneous,heinze2021conditional, gimenez2021identifying}. In the last few years, the relationship between causality, prediction, and OOD generalization has gained increasing interest since the seminal work of \cite{peters2016causal}. 
The causality-based methods rest on the long-standing assumption that causal mechanism is invariant across different domains \cite{peters2017elements}. To utilize the invariant causal mechanism and hence improve OOD generalization, some works impose restrictive assumptions on the causal diagram or structural equations \cite{subbaswamy2019preventing,heinze2021conditional,rothenhausler2021anchor,mahajan2021CausalMatching}. Another way is through recovering the causal feature \cite{rojas2018invariant,chang2020invariant, gimenez2021identifying}. For example, \cite{rojas2018invariant} proposed to select causal variables by statistical tests for equality of distributions, and \cite{chang2020invariant} leveraged some conditional independence relationships induced by the common causal mechanism assumption. It is worth noting that recovering the causal feature generally relies on restrictive assumptions, e.g., linear structural model or sufficiently many and diverse training domains \cite{rojas2018invariant,chang2020invariant,gimenez2021identifying,peters2016causal,arjovsky2019invariant,liu2021heterogeneous, krueger2021out}; see \cite{rosenfeld2020risks} for a further discussion on these two assumptions. Without these assumptions, existing methods such as invariant risk minimization \cite{arjovsky2019invariant} can fail to choose
the right predictor even in simple examples \cite{kamath2021doesIRM}. In contrast, our approach rests on a more general causal structural model and require less training domains.

\paragraph{Data Augmentation}
Data Augmentation is an important technique to the training pipeline in deep learning \cite{krizhevsky2014cifar,xie2019unsupervised,jiao2019tinybert,zhu2020freelb,zhang2018mixup,yun2019cutmix}. Commonly used methods include image rotation, cropping, Gaussian blurring, etc. With augmented data involved in training, the model generalization capability can be improved on both in-distribution \cite{shorten2019survey} and out-of-distribution data \cite{wang2021generalizing}. Different from the above mentioned works, when domain partition is available, we apply CycleGAN \cite{zhu2017unpaired} to learn the source-to-source translations to generate corresponding images with different styles, i.e., artificially generated casual invariant transformed data.
\par
Next we clarify the difference between our method and the existing methods that also involve generative models to obtain augmented data \cite{zhou2020deep,zhou2020learning,kaushikHL20}. Specifically, in \cite{zhou2020deep,zhou2020learning}, they generate data from inexistent ``novel domains'', instead of the data from known domains---{it was conjectured and empirically shown  in \cite{zhou2020learning} that source-to-source transformation could provide little help to domain generation tasks in their approach}. Roughly speaking, our augmented data represent the same ``causal feature" under different domains and we leverage the augmented data by contrasting them with the original ones, while existing methods add the average loss on the augmented data from inexistent domains to the objective. Also related is \cite{kaushikHL20} that artificially generates the counterfactually-augmented data which modify the causal feature but keep the non-causal part. Our work is different in that we use CITs to modify the non-causal feature but keep the causal part unchanged. Moreover, in our experiments, these CITs are obtained from prior knowledge or learned from training data, without the need of \emph{human manipulations to each training datum}.

\section{OOD Generalization via Causality}
In this section, we consider a general causal structural model for the OOD generalization problem. We prove that the minimax optimality of a model can be obtained via the causal feature, even if we only have access to the data from a single domain. However, as discussed in the introduction, it may be hard to recover the causal feature exactly. We hence proceed with the aid of CITs and show that a learned model can achieve the same guaranteed OOD performance.
\subsection{Invariant Causal Mechanism}
\label{sec:icm3}
We begin with a formal definition of the causal structural model used in this paper. In practice, data distributions can vary across domains, but the causal mechanism usually remains unchanged \cite{peters2017elements}. We consider the following  causal structural model to describe the data generating mechanism:
\begin{equation}\label{eq: structural model}
Y = m(g(X),\eta),\ \eta \Perp g(X)\ \text{and}\ \eta \sim F,
\end{equation}
where $X$, $Y$ are respectively the observed input and outcome, $g(X)$ denotes the causal feature, $\eta$ is some random noise, and $m(\cdot,\cdot)$ represents the unknown structural function. The relationship $\eta \Perp g(X)$ means that the noise $\eta$ is independent of the causal feature $g(X)$, and  $\eta\sim F$ indicates that it follows a distribution $F$ that can be unknown. 
\par

	{Notice that the structural model (\ref{eq: structural model}) imposes no assumption on the distribution of the input $X$. Thus, the distributions of the outcome $Y$ can vary with  $X$ under different environments, even though $Y$ depends on $X$ \emph{only} through the causal feature $g(X)$ in the data generating mechanism. Besides, there can be two correlations in the structural model, summarized as follows:
	\begin{enumerate}
	    \item Although the causal feature $g(X)$ is assumed to be independent of noise $\eta$, $X$ can correlate with noise $\eta$ under a certain domain. To see this, let us consider a toy example with the observed input $X = (X_{1}, X_{2})$. Here noise $\eta$ is correlated with $X_{1}$ while $\eta$ is independent of $X_{2}$. Then for the causal structural model $Y = X_{2} + \eta$, we have $g(X) = X_{2}$ and $g(X)\Perp \eta$ while the input $X$ is correlated to $\eta$.  
	    \item There may exist correlations between causal feature and other spurious features, e.g.,  correlation between the objective shape and the  image background in image classification tasks.
	\end{enumerate}
	Unlike the invariant causal mechanism, these two correlations are supposed to vary across domains and hence are called spurious correlations \cite{woodward2005making,arjovsky2019invariant}. If not treated carefully, the spurious correlations would deteriorate the performance of ERM-based machine learning methods and make the model perform poorly on the target domain \cite{shen2020stableDVD,shen2020stableSR,liu2021heterogeneous,arjovsky2019invariant}. For instance, in an image classification task involving horse and camel, it is very likely that in the training data all the horses are on the grass while the camels are in the desert. The spurious correlation between horse/camel and the background could easily mislead the model to making predictions using the background. Consequently, the trained model would be unreliable on OOD data.}
\par
Existing works consider a similar causal mechanism to (\ref{eq: structural model}) while more structural assumptions are usually imposed, e.g., $g(\cdot)$ is linear and the noise is additive  \cite{peters2016causal, pfister2019invariant, rojas2018invariant, gimenez2021identifying, liu2021heterogeneous}. 
Our structural model~(\ref{eq: structural model}) generalizes existing ones as we get rid of the two structural assumptions. Thus, our model constitutes a more flexible construction that is suitable to tasks in which the assumed linear or separable structural models appear implausible, e.g., the image classification task \cite{arjovsky2019invariant}. Besides, our algorithm proposed in \Cref{sec: RICEalgorithm} does not require \emph{explicitly} learning the causal feature $g(X)$, thus avoids dealing with the identifiability issue of $g(X)$.
\subsection{Generalization via Causal Feature}
\par
Throughout the rest of this paper, we focus on the  distributions under structural model~(\ref{eq: structural model}):
\begin{equation*}
\cP \hspace{-2pt}=\hspace{-2pt} \{P_{(X,Y)}\hspace{-1pt}\mid \hspace{-1pt}(X, Y)\hspace{-2pt}\sim\hspace{-2pt} P_{(X, Y)}~\text{under structural model~(\ref{eq: structural model})}\}\hspace{-1pt},
\end{equation*}
with fixed $g(\cdot)$, $m(\cdot,\cdot)$ and $F$. 
Our goal is to train a model that generalizes well across all distributions $P_{(X,Y)}\in\cP$ which follow the causal mechanism in structural model~(\ref{eq: structural model}). Particularly, we aim to find a model $h^*(\cdot)$ such that
\begin{equation}\label{eq: target set}
h^*(\cdot)\in \cH_{*} \coloneqq\mathop{\arg\min}_{h}\sup_{P\in \cP}\E_{P}[\cL(h(X), Y)],
\end{equation}
where $\cL(\cdot, \cdot)$ denotes a loss function, e.g., mean squared error for regression or cross entropy for classification. A similar minimax formulation appears in many existing works for the OOD generalization problem; see, e.g., \cite{rojas2018invariant,arjovsky2019invariant,liu2021heterogeneous,Bulhmann2020invariant,gimenez2021identifying}.
\par
In contrast with methods based on data from sufficiently many domains \cite{qian2019robust,rojas2018invariant,sagawa2020distributionally,liu2021heterogeneous,krueger2021out}, we next show that if $g(\cdot)$ is known, we can learn $h^*(\cdot)$ via single domain data. Let $P_{{{\rm s}}}$ be the distribution of the source domain from which the training data are collected. Denote  the set of optimal models under $P_{{\rm s}}$ based on causal feature $g(X)$ by
\begin{equation}\label{eq: causal opt}
\cH_{\rm{s}} \hspace{-1pt}=\hspace{-1pt} \left\{\phi\circ g \ \Big | \ \phi(w)\hspace{-1pt}\in\hspace{-1pt} \mathop{\arg\min}_{z} \E_{P_{{\rm s}}}[\cL(z, Y)\mid g(X) \hspace{-1pt}=\hspace{-1pt} w]\right\}\hspace{-2pt},
\end{equation}
where $\circ$ is the composition of functions. In this paper, we do not distinguish two functions of $w$ that are equal to each other ``almost surely", i.e., are equal for all $w$ except a set of probability zero. Then we have the following result.
\begin{theorem}\label{thm: invariant generalization}
	If $P_{{\rm s}} \in \cP$, then $\cH_{{\rm s}} \subseteq \cH_{*}$.
\end{theorem} 
A proof of \Cref{thm: invariant generalization} can be found in the supplementary material. \Cref{thm: invariant generalization} gives a class of models that belong to $\cH_{*}$, the set of solutions to the minimax problem defined in \cref{eq: target set}. A model in $\cH_{*}$ makes predictions via the causal feature $g(X)$, and can be learned using the single domain data if the form of $g(\cdot)$ is known.  \Cref{thm: invariant generalization} generalizes existing results in \cite{rojas2018invariant, liu2021heterogeneous},  in the sense that it is derived under a more general structural model and also readily includes more loss functions $\cL(\cdot,\cdot)$ beyond the mean squared loss and cross entropy loss considered in \cite{koyama2020out}.
\subsection{Learning via Causal Invariant Transformation}\label{subsec: OOD&CIT}
\Cref{thm: invariant generalization} shows that it is possible to use only single domain data to learn a class of optimal models $\cH_{{\rm s}}$ in the minimax sense. However, such a result requires an explicit formulation of causal feature $g(X)$, which is somehow impractical \cite{arjovsky2019invariant}. On the other hand, learning  causal mechanism from the observed data may face the issue of identifiability. Thus, in this section, we aim to learn a model of $\cH_{{\rm s}}$ without the requirement of the explicit form of $g(X)$. The idea of our method is to leverage the transformations that do not change the underlying causal feature.
\par   
Specifically, although the explicit form of $g(\cdot)$ is unknown in general, we can have prior knowledge that the causal feature should remain invariant to certain transformations $T(\cdot)$. For example, consider the horse v.s.~camel problem in \Cref{sec:icm3}. For a given image, the shape of a horse/camel could be the causal feature that determines its category. The \emph{exact} function w.r.t.~pixels representing the shape may be hard to obtain. Nevertheless, we do know that the shape does not vary with rotation or flipping. We now formally define these transformations.    
\begin{definition}[Causal Invariant Transformation (CIT)]\label{def:CIT}
	A transformation $T(\cdot)$ is called a causal invariant transformation if $(g\circ T)(\cdot) = g(\cdot)$.  
\end{definition}
Henceforth, $\cT_{g} = \{T(\cdot): (g\circ T)(\cdot) = g(\cdot)\}$ denotes the set consisting of all CITs. As shown in \Cref{lem: invariant characterize} in the supplementary material, the set $\cT_{g}$ is quite informative for $g(\cdot)$ and hence helps resolve the OOD generalization problem according to \Cref{thm: invariant generalization}. In some cases, knowing $\cT_{g}$ may be equivalent to knowing the causal feature or the causal parents of the outcome, e.g., assuming linear causal mechanism. However, in  applications like image classification, the causal relationships are complicated and the prior knowledge on CITs can be more accessible compared to that of causal parents, as illustrated at the end of \Cref{subsec: prior knowledge}.

With $\cT_{g}$, the following theorem states that $\cH_{{\rm s}}$ is available by solving a minimax problem constructed from single domain data, even for unknown $g(\cdot)$. 	
\par
\begin{theorem}\label{thm: alternative problem 1}
	If $P_{{\rm s}} \in \cP$, then for $\cH_{{\rm s}}$ defined in \cref{eq: causal opt}
	\begin{equation}\label{eq: minimax invariant problem}
	\cH_{{\rm s}} \subseteq
	\mathop{\arg\min}_{h} \sup_{T \in \cT_{g}}\E_{P_{{\rm s}}}[\cL(h(T(X)), Y)].
	\end{equation}
\end{theorem}
\par
A proof of \Cref{thm: alternative problem 1} is in the supplementary material. If the minimax problem (\ref{eq: minimax invariant problem}) has a unique minimum (when, e.g., some convexity conditions on the loss function $\cL(\cdot, \cdot)$ hold), \Cref{thm: alternative problem 1} implies that the model performs uniformly well over the transformed data obtained from the transformations in $\cT_g$ can generalize to distributions in $\cP$.
\par
Let $\cP_{{\rm aug}} = \{P_{(X^{\prime}, Y)}\mid (X, Y)\sim P_{{\rm s}},\ X^{\prime} = T(X), \ T\in \cT_{g}\}$, then we can rewrite the minimax problem in (\ref{eq: minimax invariant problem}) as 
\begin{equation}\label{eq: minimax with small range}
\mathop{\min}_{h} \sup_{P\in \cP_{{\rm aug}}}\E_{P}[\cL(h(X), Y)],
\end{equation}
Problem~(\ref{eq: minimax with small range}) has a similar form to problem~(\ref{eq: target set}). Recalling the structural model (\ref{eq: structural model}), it can be verified that $\cP_{{\rm aug}}$ is a subset of $\cP$. We then have the following two remarks: 
\begin{enumerate}
	\item $\cP_{{\rm aug}}$ can be a proper subset of $\cP$. Thus, the supremum taken in (\ref{eq: minimax with small range}) is more tractable compared with that in (\ref{eq: target set}), as we require less information of $\cP$. To see this, suppose that $(X, \eta) \sim P_{(X, \eta)} = P_{X}\times F$ for a distribution $P_{X}$ and $P_{{\rm s}} = P_{(X, m(g(X),\eta))}$. Then for any $P\in \cP_{{\rm aug}}$, we have $Y\Perp X\mid g(X)$ if $(X, Y)\sim P$. However, there can exist $P^{\prime}\in \cP$ so that $X$ is correlated with $\eta$ and hence the conditional independence no longer holds. That is,  $P^{\prime}$ lies in $\cP$ but not in $\cP_{{\rm aug}}$. 
	\item In the horse v.s.~camel example in \Cref{sec:icm3}, the spurious correlations lead to misleading supervision. The set $\cP_{{\rm aug}}$, on the other hand, is likely to contain distributions that do not have these spurious correlations or even entail opposite correlations. Thus, the model that overfits spurious correlations can not generalize well on these distributions, and can not be a solution to problem~(\ref{eq: minimax with small range}). For example, the data from some  $P\in\cP_{{\rm aug}}$ can have most horses on the desert while most camels are on grass. Thus, the model that overfits the spurious correlation between animal and background does not perform well on this distribution.  
\end{enumerate}
\par
Although \Cref{thm: alternative problem 1} provides a  way to learn a model with guaranteed OOD generalization, it may be computationally hard to calculate the supremum over $\cT_{g}$ when it contains plenty of or possibly infinite transformations. Take image classification tasks for example. Suppose that $\cT_{g}$ contains rotations of $\theta$ degree, with $\theta=1,\ldots,360$. Computing the loss over a total of $360$ transformations is computationally expensive. Thus, it is natural to ask a question: can we substitute  $\cT_{g}$ in (\ref{eq: minimax invariant problem}) with a proper subset?   
\subsection{Learning via Causal Essential Set}
\label{sec: causal_essential}
In this section, we positively answer the question at the end of \Cref{subsec: OOD&CIT}. We show that it is sufficient to use only a subset of $\cT_{g}$, referred to as \emph{causal essential set}. Next, we first give a formal definition of causal essential set and then prove that it is indeed the desired subset.
\begin{definition}[Causal Essential Set]
	\label{def:causal essential set}
	For $\cI_{g} \subseteq \cT_{g}$, $\cI_{g}$ is a causal essential set if for all $x_{1}$, $x_{2}$ satisfying $g(x_{1}) = g(x_{2})$, there are finite transformations $T_{1}(\cdot),\cdots,T_{K}(\cdot) \in \cI_{g}$ such that $(T_{1}\circ\dots\circ T_{K})(x_{1}) = x_{2}$.
\end{definition}
Clearly, there may be multiple causal essential sets, e.g., $\cT_{g}$ itself is a causal essential set. In most cases, we believe that there exists $\cI_{g}$ that is a proper subset of $\cT_{g}$. For example, rotation with one degree itself forms a causal essential set if $\cT_{g}$ is the set of rotations with $\theta=1,\ldots, 360,$ degrees.
\par
The next theorem indicates that the prior knowledge on any such causal essential set is sufficient to achieve a guaranteed OOD generalization, using only single domain data. A proof is provided in the supplementary material.
\begin{theorem}\label{thm: alternative problem 2}
	If $P_{{\rm s}} \in \cP$, then for any $\cI_{g}$ that is a causal essential set of $g(\cdot)$ and $\cH_{{\rm s}}$ defined in (\ref{eq: causal opt})
	\begin{equation}\label{eq: invariant feature problem}
	\begin{aligned}
	&\cH_{\rm s} = \mathop{\arg\min}_{h}~ \E_{P_{{\rm s}}}[\cL(h(X), Y)],\\
	&~~~~~~~~~~\text{\rm subject to}\ \ h(\cdot) = (h\circ T)(\cdot), \ \forall \ T(\cdot)\in \cI_{g}.
	\end{aligned}
	\end{equation}
\end{theorem}	
\par
Compared with \Cref{thm: alternative problem 1}, ``$\subseteq$" related to $\cH_{{\rm s}}$ in (\ref{eq: minimax invariant problem}) is replaced by ``$=$" in this theorem, which is a stronger theoretical result. Thus, one can also readily obtain the model that generalizes well on OOD data by minimizing the loss w.r.t. any data distribution induced by structural model~(\ref{eq: structural model}), but require less  CITs. In certain cases, the structure of a causal essential set is simple and is possible to be identified. Due to space limit, this is illustrated by an example in Section~\ref{app: toy example} in the supplementary material.

\subsection{Necessity of Prior Knowledge}\label{subsec: prior knowledge}
Before ending this section, we would like to clarify that the prior knowledge on causal invariant transformations can be more manageable, compared with  prior knowledge required by many existing approaches.
\par
First, assumption or prior knowledge is necessary to get guaranteed causal results from  observational data.
As previously introduced, some recent works assume \emph{a prior} particular forms of causal graphs, whose correctness cannot be tested from observational data. Notice that learning causal graphs from the data, or the so-called causal discovery methods \cite{peters2017elements,spirtes2000causation,zhu2020causal},  faces the same issue. Other works that have theoretical guarantee on OOD generalization  require  sufficiently many and diverse training domains \cite{arjovsky2019invariant} and/or incorporate prior knowledge on causal diagrams and causal mechanisms \cite{rosenfeld2020risks,heinze2021conditional,Lu2021NonlinearIRM}. Despite the valuable understanding and theoretical guarantee of OOD generalization for these causality based methods, it is not clear whether the assumed conditions or prior knowledge indeed hold in practical applications.
In general, only randomized controlled experiments are the golden standard to infer the causal relationship, summarized as ``no causation without manipulation'' \cite{holland1986statistics}. On the other hand, directly creating ``manipulated'' data can lead to a better causal effect estimation. In \cite{kaushikHL20}, authors artificially generate counterfactually-augmented data that modify the causal feature but keep the non-causal part for each sentence. Then simple processing (e.g., directly combining the observational and the manipulated data) improves the generalization performance. 
\par
These observations together motivate us to consider: \emph{how to generate manipulations to achieve a causal guarantee at a lower human cost?} For example, one need not apply human manipulations to each training datum. This consideration leads us to the CITs, on which we often have some prior knowledge. Many augmentation techniques can be used as CITs, such as rotation and Gaussian blurring. When domain partition is available, e.g., in the multi-domain learning setting \cite{peters2016causal,arjovsky2019invariant}, the CITs can be learned from the data. For example, we  employ  GAN or other generative models to synthesize ``manipulated'' data in \Cref{subsec: generalizing on unseen domains}. We believe that such prior knowledge on CITs is more accessible than artificially manipulated data, and is manageable in practical applications. 
\par
\begin{figure}[t!]
\centering
\begin{subfigure}[t]{0.25\linewidth}
    \centering
    \includegraphics[scale = 0.5]{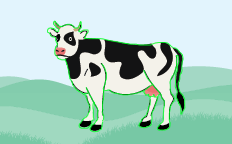}
\end{subfigure} 
\hspace{-0.2in}
\begin{subfigure}[t]{0.25\linewidth}
     \centering
    \includegraphics[scale = 0.5]{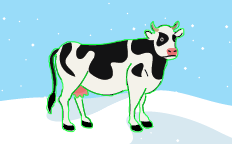}
\end{subfigure}
\hspace{-0.2in}
\begin{subfigure}[t]{0.25\linewidth}
     \centering
    \includegraphics[scale = 0.5]{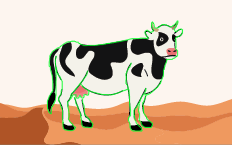}
\end{subfigure}
    \caption{Profiles of cows in these figures are considered as causal features, while the different backgrounds are spurious features.}\label{fig: cow}
    \vspace{-0.15in}
\end{figure}

Finally, we clarify that prior knowledge on CITs can also be more accessible than that of causal feature in many tasks, as illustrated by \Cref{fig: cow} from \cite{cloudera2021causality}. In \Cref{fig: cow}, profiles of cows (highlighted by green lines) are considered as causal feature, and a prediction model \emph{only} based on such causal feature can generalize well to images with different backgrounds. However, we do not know which pixels or what function of pixels represent the profile; indeed, the profile depends on different pixels in the first and the second images in \Cref{fig: cow}. In contrast, we readily know that changing the background of the image, e.g., transforming the second image to the third, will not affect the causal feature. This can be treated as our prior knowledge on CITs.

\section{Algorithm}
\label{sec: RICEalgorithm}
We now propose an algorithm based on the previous analysis w.r.t.~CITs. Let $D(\cdot,\cdot)$ denote some measure of discrepancy satisfying $D(v_{1}, v_{2}) = 0$ if $v_{1} = v_{2}$ and $D(v_{1}, v_{2}) > 0$ otherwise. Then for any model $h(\cdot)$ and transformation $T(\cdot)$, $\E_{P_{{\rm s}}}[D\big(h(X), h(T(X))\big)] = 0$ implies $h(\cdot) = h(T(\cdot))$ almost surely. Together with \Cref{thm: alternative problem 2}, we consider the following formulation
\begin{equation*}
\begin{aligned}
\min_{h}&~~\E_{P_{{\rm s}}}\left[\cL(h(X), Y)\right],\quad\\ 
\textrm{subject to}&~~\E_{P_{{\rm s}}}\left[\sup_{T\in \cI_{g}}D\big(h(X), h(T(X))\big)\right] = 0,
\end{aligned}
\end{equation*}
where $\cI_{g}$ is a causal essential set. To obviate the difficulty of solving a constrained optimization problem, we further consider to minimize a regularized formulation
\begin{equation} \label{eq: population penalize}
\begin{aligned}
\E_{P_{{\rm s}}} [\cL(h(X), Y)] \!+\! \lambda_{0}\E_{P_{{\rm s}}}\bigg[\sup_{T\in \cI_{g}}[D(h(X), h(T(X)))\bigg]
\end{aligned}
\end{equation}
with a given regularization constant $\lambda_{0}>0$. Supposing  that we have training samples $\{(x_{i}, y_{i})\}_{i=1}^{n}$, then we propose to minimize the  empirical counterpart of (\ref{eq: population penalize})
\begin{equation}
\begin{aligned}
\!\frac{1}{n}\!\sum\limits_{i=1}^{n}\!\cL(h(x_{i}), y_{i})\!+ \! \frac{\lambda_{0}}{n}\sum\limits_{i=1}^{n}\!\sup_{T\in \cI_{g}}\![D(h(x_{i}), h(T(x_{i})))].
\end{aligned}\nonumber
\end{equation} 
We then propose Algorithm \ref{alg: RICE}, Regularized training with Invariance on Causal Essential set (RICE), to solve the above problem, where the update step in line $7$ can be substituted by other optimization algorithms, e.g., Adam \cite{kingma2015adam}.
\par
Note that obtaining a complete causal essential set may also be hard in many applications. Nonetheless, we usually have or can learn certain transformations with the desired causal invariance. We will empirically show that the proposed RICE enables an improved OOD generalization, even with a set of only a few CITs. In this case, we can simply replace $\cI_g$ with this set in Algorithm \ref{alg: RICE}.

\begin{algorithm}[t!]
	\caption{Regularized training with Invariance on Causal Essential set (RICE).}
	\label{alg: RICE}
	\textbf{Input:} Training set $\{(x_{1}, y_{1}), \cdots, (x_{n}, y_{n})\}$, batch size $S$, learning rate $\eta$,  training iterations $N$, model $h_{\beta}(\cdot)$ with parameter $\beta$, initialized parameter $\beta_{0}$, regularization constant $\lambda_{0}$, causal essential set $\cI_{g}$, and discrepancy measure $D(\cdot, \cdot)$.
	\begin{algorithmic}[1]
		\For 	{$i=1, \ldots, n$ }  
		\State{Generate transformed samples $\{T(x_{i})\}_{T\in \cI_{g}}$.}
		\EndFor
		\For        {$t=0, \cdots ,N$} 
		\State  {Randomly sample a mini-batch $\cS = \{(x_{t_{1}}, y_{t_{1}}),$ $\cdots, (x_{t_{S}}, y_{t_{S}})\}$ from training set.
			\State Fetch the transformed samples $\{T(x_{t_1})\}_{T\in \cI_{g}},$ $\cdots,$ $\{T(x_{t_{S}})\}_{T\in \cI_{g}}$.}		
		\State  {Update model parameters via first-order method e.g., stochastic gradient descent:}
		\vspace{-0.1in}
		 {\small
			\[\begin{aligned}
			\beta_{t + 1} &= \beta_{t} - \frac{\eta}{S}\sum\limits_{i=1}^{S}\nabla_{\beta}\cL(h_{\beta}(x_{t_{i}}), y_{t_{i}})\Big|_{\beta = \beta_{t}} + \eta\nabla_{\beta}\Big\{\frac{\lambda_{0}}{S}\sum\limits_{i=1}^{S}\sup\limits_{T\in \cI_{g}}D\big(h_{\beta}(x_{t_{i}}), h_{\beta}(T(x_{t_{i}}))\big)\Big\}\Big|_{\beta = \beta_{t}}\hspace{-1pt}.
			\end{aligned}\]}
			\vspace{-0.1in}
		\EndFor
	\end{algorithmic}
\end{algorithm}

\section{Experiments}\label{sec:experiment}
In this section, we empirically evaluate the efficacy of the proposed algorithm RICE on real-world datasets. We train the model using data from some of the available domains and evaluate the performance on the data from the rest domains that are not used in training. 
As suggested in \cite{ye2021ood}, the OOD data can be classified into two categories, namely, data with \emph{correlation shift} or with \emph{diversity shift}. 
Empirical results show that RICE can handle both kinds of OOD data. Due to space limit, part of results, including a toy experiment of synthetic data mentioned in \Cref{sec: causal_essential} and ablation studies, are given in the supplementary material.\footnote{Part of the experiments was  supported by MindSpore (\url{https://www.mindspore.cn}), a deep learning computing framework.}

\begin{figure*}[h]
	\centering
	\includegraphics[scale=0.48]{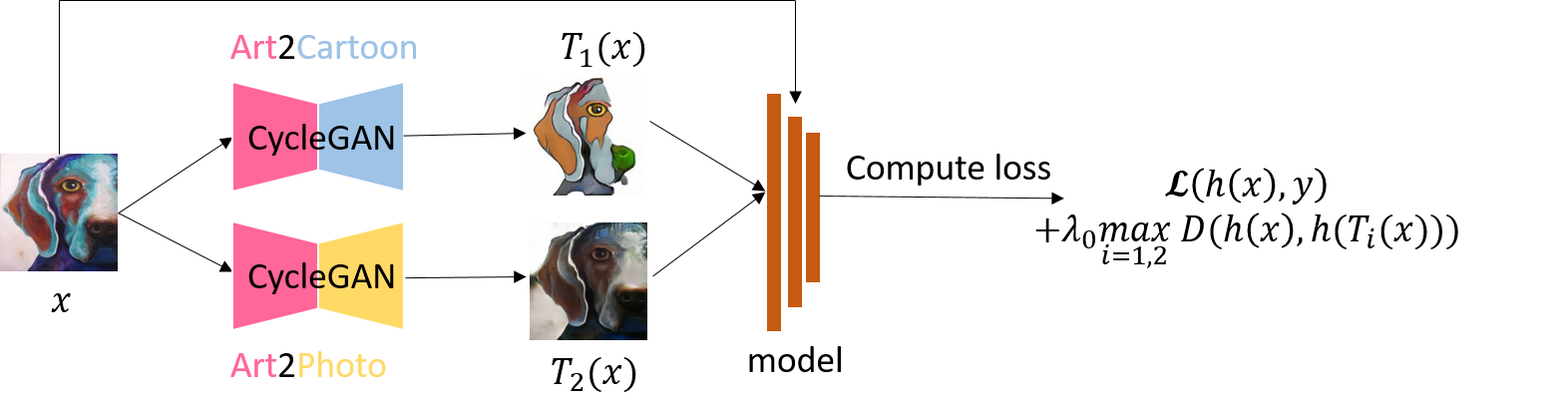}
 	\vspace{-0.1in}
	\caption{The proposed algorithm RICE on the \texttt{PACS} dataset. The training data are from domains \{art, cartoon, photo\}, and we want the model to perform well on the sketch data. This figure describes the training procedure of RICE for a training image from the art domain.}
	\label{fig: algorithm}
\end{figure*}
\subsection{Breaking Spurious Correlation}\label{subsec: spurious correlation}
As we have discussed in \Cref{sec:icm3}, the spurious correlation in the data may mislead the model to  wrong predictions on OOD data, resulting in correlation shift. In this section, we empirically verify that RICE in Algorithm \ref{alg: RICE} can obviate overfitting such spurious correlations. 
\vspace{-0.5em}
\paragraph{Data} We use the colored \texttt{MNIST} (\texttt{C-MNIST}) dataset from \cite{arpit2019predicting}. As in \cite{arpit2019predicting}, we vary the colors of both foreground and background of an image.
\par
The original \texttt{MNIST} dataset consists of handwritten digits from ten categories, namely, $0$ to $9$. To construct a training set of \texttt{C-MNIST}, we pick two colors for the foreground of the images in a given category, and then randomly replace the foreground color with one of the two colors assigned to the category. The background color of each image is handled similarly. For the test set, we randomly assign colors to the foreground and background of each image from the \texttt{MNIST} test set, regardless of its category. Some images from the generated \texttt{C-MNIST} dataset are visualized in \Cref{fig: c-mnist} in the supplementary material. The construction introduces a spurious correlation between category and color in the training set, but not in the test set. In the following, we will show that the proposed method RICE will not be affected by this spurious correlation. 
\vspace{-0.5em}
\paragraph{Setup} Our model is a five-layer convolution neural network as in \cite{arpit2019predicting}. For RICE, we choose the cross entropy loss for  $\cL(\cdot, \cdot)$ and the $\ell_{2}$-distance for $D(\cdot, \cdot)$. The model is updated by Adam \cite{kingma2015adam}, and other hyperparameters are given in the supplementary material. For a handwritten digit, it is known that the shape of its foreground, rather than the color of either foreground or background, determines its category. Thus, transforming the image background with a  color (e.g., black) and its foreground with another color (e.g., white) would be a desired CIT. In our experiment, we simply use the original \texttt{MNIST} images as the transformed data to show the effectiveness of the proposed method.
\par 
As we use the original \texttt{MNIST} dataset in training,  the training data can be seen from two domains, i.e., the original \texttt{MNIST} and the \texttt{C-MNIST} datasets. As such, we compare RICE  with several widely used domain generalization algorithms using  the same training data, including empirical risk minimization (ERM), ERM with Mixup \cite[Mixup]{zhang2018mixup}, marginal transfer learning \cite[MTL]{blanchard2021domain}, group distributionally robust optimization \cite[GroupDRO]{sagawa2020distributionally}, domain-adversarial neural networks \cite[DANN]{ganin2016domain} and invariant risk minimization \cite[IRM]{arjovsky2019invariant}.  See the supplementary material for a further introduction of these algorithms. For these baseline algorithms, the hyperparameters are adopted from \cite{gulrajani2020search}. We remark that  the same training data, i.e., \texttt{MNIST} and \texttt{C-MNIST}, are used in our method and baseline methods 

\begin{table}[t!]
	\caption{Accuracy (\%) on the \texttt{C-MNIST} test set.}
	\vspace{-0.1in}
	\label{tbl:c-mnist}
	\setlength\tabcolsep{2pt}
	\centering
	\scalebox{0.8}{
		{
			\begin{tabular}{l|cccccccccc}
				\hline
				Dataset   &  ERM    &   Mixup  & MTL    &   GroupDRO   &   DANN   &   IRM  &   RICE(OURS)   \\
				\hline
				\texttt{C-MNIST}   &  13.3  &  17.5  &  14.7  &   14.1  &   28.1   &    15.8   & \textbf{96.9} \\
				\hline
	\end{tabular}}}
	\vspace{-0.15in}
\end{table}

\vspace{-0.5em}
\paragraph{Main Results} The empirical results are reported in  \Cref{tbl:c-mnist}. We observe that only the proposed algorithm RICE  works well on the OOD data in this experiment. We speculate that this is because, for the baseline algorithms, the misleading supervision signal from the color is memorized by the models, even though the original \texttt{MNIST} images are also included in the training set. However, for RICE, the regularizer penalizes the discrepancy between the model outputs of the colored images and the corresponding \texttt{MNIST} versions, which makes the model insensible to the spurious correlation but more dependent on the invariant causal feature.
\subsection{Generalizing to Unseen Domains}\label{subsec: generalizing on unseen domains}
In this section, we conduct experiments on two benchmark datasets, \texttt{PACS} and \texttt{VLCS}, which are commonly used in domain generalization. The two datasets correspond to the diversity shift that we have mentioned. As in other related works in domain generalization \cite{zhou2020deep,zhou2021domain,gulrajani2020search}, the domain labels are known in the training set.
\vspace{-0.5em}
\paragraph{Data} \texttt{PACS} is an image classification dataset consisting of data from four domains of different styles, i.e., \{art, cartoon, photo, sketch\}, with seven different categories in each domain. \texttt{VLCS} is a dataset comprised of four photographic domains: \{VOC2007, LabelMe, Caltech101, SUN09\}, and each domain contains five different categories. 
\vspace{-0.5em}
\paragraph{Setup} As in \cite{gulrajani2020search}, we use the model ResNet50 \cite{he2016deep} pre-trained on \texttt{ImageNet} \cite{deng2009imagenet} as the backbone model, and fine-tune the model with different baseline methods. For RICE, the model is trained by Adam and the used hyperparameters are provided in the supplementary material. To implement RICE, we need to generate  casually invariant transformed data. In the \texttt{PACS} dataset, each domain represents a style of images, e.g., photo or art. Since varying the style of an image does not change its category, we construct the transformations that modify image styles as CITs. To this end, we use CycleGAN \cite{zhu2017unpaired} to learn the transformations for each pair of domains in the training set, and then implement  RICE  using the trained CycleGAN models. In \texttt{VLCS}, the photographic of the image plays a similar role to the style in \texttt{PACS}, and we also apply CycleGAN to learning the transformations. Other generative models may also be used, e.g., StarGAN \cite{choi2018stargan}, for data from numerous domains.
\par
The procedure of RICE is summarized in \Cref{fig: algorithm}. We also compare the proposed RICE with other commonly used domain generalization algorithms, as in the previous experiment. For better results and a fair comparison with baseline methods, we also provide an ablation study with single domain training data in the supplementary material. 
\vspace{-0.5em}
\paragraph{Main Results}
The experimental results on \texttt{PACS} and \texttt{VLCS} are summarized in Tables \ref{tbl:PACS} and \ref{tbl:VLCS}, respectively. The results of baseline methods are from \cite{gulrajani2020search}. The proposed RICE exhibits better OOD generalization capability compared with the baseline methods on the \texttt{PACS} and \texttt{VLCS} datasets, in terms of the average and particularly the worst-case test accuracies. 
Here we provide an intuitive explanation to the performance of RICE using \texttt{PACS} as an example. From \Cref{fig: pacs} in the supplementary material, we can see that the trained  CycleGAN model is likely to introduce spurious correlation related to domains, and models which capture the spurious correlation would be penalized in RICE. RICE can achieve an improved performance because it seeks to make predictions on top of the causal feature (e.g., shape of the object in the images), rather than the spurious feature related to the domains (e.g., style for \texttt{PACS}). Moreover, as also seen from \Cref{fig: pacs}, some generated images from CycleGAN are indeed not similar to the original images and are also blurring, which indicates that our RICE is robust to the quality of generated data. 
\par
Finally, we verify whether the improved performance of RICE originates solely from the augmented data generated by CycleGAN. We conduct ERM training with these augmented data on \texttt{PACS}. The test accuracies on domains P, A, C, S are respectively 96.2, 84.9, 81.2, 80.5, which are worse than those of RICE. We believe that such a performance, which is consistent with previous observations in \cite{zhou2020learning}, demonstrates that the regularizer  that compares different augmentations with the original data is critical to capture the invariance and to improve the model robustness.
\begin{table}[t!]
	\caption{Test accuracy (\%) of ResNet50 on the \texttt{PACS} dataset.}
    \vspace{-0.1in}
	\label{tbl:PACS}
	\centering
	\scalebox{0.85}{
		{
			\begin{tabular}{c|*{7}{c}}
				\hline
				Method & P & A & C & S & Avg & Min \\
				\hline
				ERM & 97.2 & 84.7 & 80.8 & 79.3 & 85.5 & 79.3 \\
				Mixup & 97.6 & 86.1 & 78.9 & 75.8 & 84.6 & 75.8 \\
				MTL & 96.4 & 87.5 & 77.1 & 77.3 & 84.6 & 77.1 \\
				GroupDRO & 96.7 & 83.5 & 79.1 & 78.3 & 84.4 & 78.3\\
				DANN & 97.3 & 86.4 & 77.4 & 73.5 & 83.6 & 73.5 \\
				IRM & 96.7 & 84.8 & 76.4 & 76.1 & 83.5 & 76.1 \\
				RICE (OURS) & 96.8 & 87.8 & 84.3 & 84.7 & \bf{88.4} & \bf{84.3}\\
				\hline
	\end{tabular}}}
\end{table}

\begin{table}[t!]
	\caption{Test accuracy (\%) of ResNet50 on the \texttt{VLCS} dataset.}
	\vspace{-0.1in}
	\label{tbl:VLCS}
	\centering
	\scalebox{0.85}{
		{
			\begin{tabular}{c|*{7}{c}}
				\hline
				Method & V & L & C & S & Avg & Min\\
				\hline
				ERM & 74.6 & 64.3 & 97.7 & 73.4 & 77.5 & 64.3\\
				Mixup & 76.1 & 63.4 & 98.4 & 72.9 & 77.7 & 63.4 \\
				MTL & 75.3 & 64.3 & 97.8 & 71.5 & 77.2 & 64.3\\
				GroupDRO & 76.7 & 63.4 & 97.3 & 69.5 & 76.7 & 63.4\\
				DANN & 77.2 & 65.1 & 99.0 & 73.1 & 78.6 & 65.1\\
				IRM & 77.3 & 64.9 & 98.6 & 73.4 & 78.5 & 64.9\\
				RICE (OURS) & 75.1 & 69.2 & 98.3 & 74.6 & \bf{79.3} & \bf{69.2}\\
				\hline
	\end{tabular}}}
	\vspace{-0.1in}
\end{table}

\section{Concluding Remarks}
In this paper, we theoretically show that knowledge of the CITs makes it feasible to learn an OOD generalized model via single domain data. The CITs can be either obtained from prior knowledge or learned from training data, without the need of human manipulations to each training datum. Inspired by our theoretical findings, we propose RICE to achieve an enhanced OOD generalization capability and the effectiveness of RICE is demonstrated empirically over various experiments. 
\par
In our experiments, we focus on image classification tasks for domain generalization. Nevertheless, our theory and the proposed algorithm can apply to other datasets once some CITs are available. For example, in natural language processing (NLP), changing the position of the adverbial or synonym substitution does not change the semantic meaning and hence can be treated as CITs. However, generating such causally invariant sentences via deep learning method is not easy. How to ease the generation and apply our RICE to NLP domain is left as a future work.

\small{
\bibliographystyle{ieee_fullname}
\bibliography{ISM-CVPR}}
\appendix
\include{appendix}
\clearpage

\end{document}

%% file: math_commands.tex
\usepackage{amsmath,amsfonts,bm}

\def\eqref#1{equation~\ref{#1}}

\def\1{\bm{1}}

\DeclareMathAlphabet{\mathsfit}{\encodingdefault}{\sfdefault}{m}{sl}
\SetMathAlphabet{\mathsfit}{bold}{\encodingdefault}{\sfdefault}{bx}{n}

\def\cC{{\mathcal{C}}}
\def\cD{{\mathcal{D}}}

\def\cH{{\mathcal{H}}}
\def\cI{{\mathcal{I}}}

\def\cL{{\mathcal{L}}}

\def\cP{{\mathcal{P}}}

\def\cR{{\mathcal{R}}}
\def\cS{{\mathcal{S}}}
\def\cT{{\mathcal{T}}}
\def\cU{{\mathcal{U}}}

\def\T{{\rm T}}

\newcommand{\E}{\mathbb{E}}



%% file: appendix.tex
\onecolumn
\section{Proofs}\label{app: proofs}
\subsection{Proof of Theorem \ref{thm: invariant generalization}}\label{app: proof of thm i. g.}
\paragraph{Restatement of Theorem \ref{thm: invariant generalization}}
\emph{If $P_{{\rm s}} \in \cP$, then $\cH_{{\rm s}} \subset \cH_{*}$.}
\begin{proof}
	It suffices to prove that for any $h_{{\rm s}} \in \cH_{{\rm s}} $, we have
	\begin{equation}\label{eq: single hypo opt}
	h_{{\rm s}}(\cdot) \in \mathop{\arg\min}_{h} \sup_{P\in \cP}\E_{P}[\cL(h(X), Y)].
	\end{equation}
	To prove (\ref{eq: single hypo opt}), we only need to show that for any $h(\cdot)$ and $P \in \cP$, there exists $Q \in \cP$ such that 
	\begin{equation}\label{eq: sufficient eq 1}
	\E_{Q}[\cL(h(X), Y)] \geq \E_{P}[h_{{\rm s}}(X), Y)],
	\end{equation}
	and hence
	\begin{equation}
	\sup_{Q\in \cP}\E_{Q}[\cL(h(X), Y)] \geq \sup_{P\in \cP}\E_{P}[\cL(h_{{\rm s}}(X), Y)]. \nonumber
	\end{equation}
	Recall that 
	\begin{equation}
	\cH_{\rm{s}} = \left\{(\phi\circ g)(\cdot)\ \Big | \ \phi(w) \in \mathop{\arg\min}_{z} \E_{P_{{\rm s}}}[\cL(z, Y)\mid g(X) = w], \quad a.s.\right\}.\nonumber
	\end{equation}
	Since $h_{{\rm s}}(\cdot) \in \cH_{{\rm s}}$, there is some $\phi_{{\rm s}}(\cdot)$ satisfying $h_{{\rm s}}(\cdot) = (\phi_{{\rm s}}\circ g)(\cdot)$ and $\phi_{{\rm s}}(w) \in \mathop{\arg\min}_{z} \E_{P_{{\rm s}}}[\cL(z, Y)\mid g(X) = w]$ for almost every $w$.
	Suppose $(X, \eta) \sim P_{X}\times F$ and $(m(g(X), \eta), X)\sim Q$ where $P_{X}$ is the marginal distributions of $X$ under $P$. Let $\cU$ be the support of noise $\eta$. Then 
	\begin{equation}
	\begin{aligned}
	\E_{Q}[\cL(h(X), Y)\mid X = x] & = \int_{\cU} \cL(h(x), m(g(x), u))P_{\eta}(du)\\
	& \geq \int_{\cU} \cL(\phi_{{\rm s}}(g(x)), m(g(x), u))P_{\eta}(du) \\
	& = \E_{P}[\cL(h_{{\rm s}}(X), Y)\mid g(X) = g(x)] \quad a.s. \nonumber
	\end{aligned}
	\end{equation}
	Here the first equation follows from the fact that $X$ and $\eta$ are independent under $Q$. The inequality is from the fact that
	\[\int_{\mathcal{U}} \mathcal{L}(h(x), m(g(x), u))P_{\eta}(du) = \mathbb{E}_{P_{\rm s}}[\mathcal{L}(h(x), Y) \mid g(X) = g(x)]\]
	and
	\[\phi_{{\rm s}}(w) \in \mathop{\arg\min}_{z} \E_{P_{{\rm s}}}[\cL(z, Y)\mid g(X) = w] = \mathop{\arg\min}_{z} \int_{\cU}\cL(z, m(w, u))P_{\eta}(du)\]
	for almost every $w$. The last equation is due to $P \in \cP$.
	Then equation (\ref{eq: sufficient eq 1}) follows by taking expectation and the law of iterated expectation.
\end{proof}

\subsection{Proof of Theorem \ref{thm: alternative problem 1}}\label{app: proof of thm minimax and restriction}
To begin with, we establish two useful lemmas regarding CITs. The first lemma states that $g(\cdot)$ is determined up to an invertible transformation by the transformation that it is invariant to. 

For a given function $h(\cdot)$, let $\cT_{h} = \{T(\cdot): (h\circ T)(\cdot) = h(\cdot)\}$. Then we have the following  lemma.
\begin{lemma}\label{lem: invariant characterize}
	For any $h_{1}(\cdot)$ and $h_{2}(\cdot)$,
	$\cT_{h_{1}} \subset  \cT_{h_{2}}$ if and only if there exists a function $v(\cdot)$ such that $h_{2}(\cdot) = (v \circ h_{1})(\cdot)$, and $\cT_{h_{1}} = \cT_{h_{2}}$ if and only if there is an invertible function $v(\cdot)$ such that $h_{2}(\cdot) = (v \circ h_{1})(\cdot)$.
\end{lemma}
\par
\begin{proof}
	We only prove the former statement as the latter can be obtained as a corollary of the former. The ``if" direction is obvious. 
	\par
	Here we prove the ``only if" direction. Let $\cR_{1}$ and $\cR_{2}$ be the range of $h_{1}(\cdot)$ and $h_{2}(\cdot)$, respectively. For any $w_{1}\in \cR_{1}$ and $w_{2} \in \cR_{2}$, define $\cD_{h_{1},w_{1}} = \{x: h_{1}(x) = w_{1}\}$ and $\cD_{h_{2},w_{2}} = \{x: h_{2}(x) = w_{2}\}$. Then $h_{2}(\cdot) = (v \circ h_{1})(\cdot)$ if and only if for any $w_{2} \in \cR_{2}$, there is some $w_{1}\in\cR_{1}$ such that $\cD_{h_{1},w_{1}} \subset \cD_{h_{2},w_{2}}$. Thus, the former claim holds if we can show the following: $\cT_{h_{1}} \subset \cT_{h_{2}}$ implies that there is some $w_{2}\in\cR_{2}$ such that $\cD_{h_{1},w_{1}} \subset \cD_{h_{2},w_{2}}$ for any $w_{1} \in \cR_{1}$. We will prove this by contraction. 
	\par
	Suppose there exists $w_{1}$ such that $\cD_{h_{1},w_{1}} \not\subset \cD_{h_{2},w_{2}}$ for any $w_{2}\in\cR_{2}$. Because $\bigcup_{w_{2}\in\cR_{2}}\cD_{h_{2},w_{2}}$ constitutes the whole space, there is some $w_{2}$ such that $\cD_{h_{1}, w_{1}}\bigcap\cD_{h_{2},w_{2}} \neq \varnothing$ and $\cD_{h_{1}, w_{1}}\not\subset\cD_{h_{2},w_{2}}$. Thus,  $\cD_{h_{1}, w_{1}}\setminus \cD_{h_{2},w_{2}} \neq \varnothing$. Let $x^{\dag}$ denote a point in
	$\cD_{h_{1},w_{1}}\setminus \cD_{h_{2},w_{2}}$ and let $x^{\prime}$ a point in $\cD_{h_{2},w_{2}}\bigcap\cD_{h_{1},w_{1}}$. Define $T_{*}$ as the transformation such that $T_{*}(x^{\prime}) = x^{\dag}$, $T_{*}(x^{\dag}) = x^{\prime}$ and $T_{*}(x) = x$ for $x \neq \{x^{\prime}, x^{\dag}\}$. Then it is straightforward to verify that $T_{*}\in \cT_{h_{1}}$ but $T_{*}\notin \cT_{h_{2}}$, which is a contradiction.
\end{proof}
Thus $g(\cdot)$ can be characterized by $\cT_{g}$ up to an invertible transformation. Define $\cC_{g} = \{g^{\prime}(\cdot): g^{\prime}(\cdot) = (v\circ g)(\cdot) \ \text{for some invertible transformation $v(\cdot)$}\}$. For any $g^{\prime} \in \cC_{g}$, by defining 
\begin{equation}
\cH_{\rm{s}}^{\prime} = \left\{(\phi\circ g)(\cdot)\ \Big | \ \phi(w) \in \mathop{\arg\min}_{z} \E_{P_{{\rm s}}}[\cL(z, Y)\mid g^{\prime}(X) = w], \quad a.s.\right\}, \nonumber
\end{equation}
similar arguments as in the proof of Theorem \ref{thm: invariant generalization} can show $\cH_{{\rm s}}^{\prime} \subset \cH_{*}$.
To train a model that generalizes well on all the data distributions following the same causal mechanism, any $g^{\prime}(\cdot)\in \cC_{g}$ is sufficient. Thus, if $\cT_{g}$ is known, to find a model belongs to $\cH_{*}$, one may firstly find an invariant feature map $g^{\prime}(\cdot)$ such that $\cT_{g^{\prime}} = \cT_{g}$ and then obtain the model according to Theorem \ref{thm: invariant generalization}. However, finding a $g^{\prime}(\cdot)$ such that $\cT_{g^{\prime}} = \cT_{g}$ is sometimes still a hard task.

For any function $h(\cdot)$,  define $\cI_{h}$ in the same way as $\cI_{g}$ with $g(\cdot)$ replaced by $h(\cdot)$ in the definition. We then have the following lemma.
\begin{lemma}\label{lem: essential invariant set}
	For any $h_{1}(\cdot)$ and $h_{2}(\cdot)$, if $\cI_{h_{1}} \subset \cT_{h_{2}}$, then there exists a function $v(\cdot)$ such that $h_{2}(\cdot) = (v \circ h_{1})(\cdot)$.
\end{lemma}
\par
\begin{proof}
	Like in the proof of Lemma (\ref{lem: invariant characterize}), it suffices to show that $\cI_{h_{1}} \subset \cT_{h_{2}}$ implies for any $w_{1} \in \cR_{1}$, there is some $w_{2}\in\cR_{2}$ such that $\cD_{h_{1},w_{1}} \subset \cD_{h_{2},w_{2}}$. We prove this by contraction. 
	
	Suppose there is some $w_{1}$ such that $\cD_{h_{1},w_{1}} \not\subset \cD_{h_{2},w_{2}}$ for any $w_{2}\in\cR_{2}$. Because $\bigcup_{w_{2}\in\cR_{2}}\cD_{h_{2},w_{2}}$ is the whole space, there is some $w_{2}$ such that $\cD_{h_{1},w_{1}}\bigcap\cD_{h_{2},w_{2}} \neq \emptyset$ and $\cD_{h_{1},w_{1}}\not\subset\cD_{h_{2},w_{2}}$. Thus,  $\cD_{h_{1},w_{1}}\setminus \cD_{h_{2},w_{2}} \neq \emptyset$. Let $x^{\dag}$ be a point in
	$\cD_{h_{1},w_{1}}\setminus \cD_{h_{2},w_{2}}$ and let $x^{\prime}$ be a point in $\cD_{h_{2},w_{2}}\bigcap\cD_{h_{1},w_{1}}$. According to the definition of essential invariant subset, because $h_{1}(x_{1}) = h_{2}(x_{2})$, there are finite transformations $T_{1}(\cdot),\dots,T_{K}(\cdot) \in \cI_{g}$ such that $\bar{T}(x^{\prime}) = x^{\dag}$ where $\bar{T}(\cdot) = (T_{1}\circ\dots\circ T_{K})(\cdot)$. It can be verified that $\cT_{h_{2}}$ is closed with respect to function composition. Hence, $\bar{T}(\cdot) \in \cT_{h_{2}}$. However, $h_{2}(\bar{T}(x^{\prime})) = h_{2}(x^{\dag}) \neq w_{2} = h_{2}(x^{\prime})$, which is a contradiction. 
\end{proof}
\paragraph{Restatement of Theorem \ref{thm: alternative problem 1}}
\emph{If $P_{{\rm s}} \in \cP$, then
	\begin{equation}
	\cH_{{\rm s}} \subset
	\mathop{\arg\min}_{h} \sup_{T \in \cT_{g}}\E_{P_{{\rm s}}}[\cL(h(T(X)), Y)],\nonumber
	\end{equation}
	where $\cH_{{\rm s}}$ is defined in (\ref{eq: causal opt}).
}
\begin{proof}
	It suffices to show that for all $h_{{\rm s}}(\cdot)\in \cH_{{\rm s}}$, we have
	\begin{equation}
	h_{{\rm s}}(\cdot) \in
	\mathop{\arg\min}_{h} \sup_{T \in \cT_{g}}\E_{P_{{\rm s}}}[\cL(h(T(X)), Y)].
	\end{equation} 
	Note that $h_{{\rm s}}(\cdot) = (\phi_{{\rm s}}\circ g)(\cdot)$ for some $\phi_{{\rm s}}(\cdot)$ and hence is invariant to any transformation $T(\cdot)\in \cT_{g}$. We then have $\sup_{T \in \cT_{g}}\E_{P_{{\rm s}}}[\cL(h_{{\rm s}}(X), Y)] = \E_{P_{{\rm s}}}[\cL(h_{{\rm s}}(T(X)), Y)]$. Thus, it suffices to prove that for all $h(\cdot) $, there exists $T(\cdot) \in \cT_{g}$ such that 
	\begin{equation}\label{eq: sufficient eq 2}
	\E_{P_{{\rm s}}}[\cL(h(T(X)), Y)] \geq \E_{P_{{\rm s}}}[\cL(h_{{\rm s}}(X), Y)].
	\end{equation}
	According to axiom of choice, there is a choice function $a$ such that $a(w) \in \cD_{g,w}$ for almost every $w$. Define $\tilde{T}$ to be a transformation such that $\tilde{T}(x) = a(w)$ for $x\in \cD_{g,w}$. Then $\tilde{T}(\cdot) \in \cT_{g}$ and we have
	\begin{equation}
	\begin{aligned}
	\E_{P_{{\rm s}}}[\cL(h(\tilde{T}(X)), Y)\mid g(X) = w] & = \E_{P_{{\rm s}}}[\cL(h(a(w)), Y)\mid g(X) = w] \\
	& \geq  \E_{P_{{\rm s}}}[\phi_{{\rm s}}(w), Y)\mid g(X) = w] \\
	& = \E_{P_{{\rm s}}}[\cL(h_{{\rm s}}(X), Y)\mid g(X) = w] \quad a.s. 
	\end{aligned}
	\end{equation}
	By taking expectation on both sides, we can obtain equation~(\ref{eq: sufficient eq 2}).
\end{proof}

\section{Proof of Theorem \ref{thm: alternative problem 2}}\label{app: proof of thm: alternative problem 2}
\paragraph{Restatement of Theorem \ref{thm: alternative problem 2}}
\emph{If $P_{{\rm s}} \in \cP$,
	then 
	\begin{equation}\label{eq: invariant feature problem-copy}
	\begin{aligned}
	\cH_{\rm s} = \mathop{\arg\min}_{h} \E_{P_{{\rm s}}}[\cL(h(X), Y)]\quad
	\text{\rm subject to}\ \ h(\cdot) = (h\circ T)(\cdot), \ \forall \ T(\cdot)\in \cI_{g}.
	\end{aligned}
	\end{equation}
	where $\cI_{g}$ is any causal essential set of $g(\cdot)$ and $\cH_{{\rm s}}$ is defined in (\ref{eq: causal opt}).}

\begin{proof}
	We first show
	\begin{equation}
	\begin{aligned}
	\cH_{\rm s} \subset &\mathop{\arg\min}_{h} \E_{P_{{\rm s}}}[\cL(h(X), Y)]\\
	&~\text{subject to}~\ \ h(\cdot) = (h\circ T)(\cdot), \ \forall \ T(\cdot)\in \cI_{g}.\nonumber
	\end{aligned}
	\end{equation}
	Note that the restriction in (\ref{eq: invariant feature problem-copy}) is equivalent to $\cI_{g} \subset \cT_{h}$.
	It suffices to show that 
	\begin{equation}\label{eq: proof-constrained opt}
	\E_{P_{{\rm s}}}[\cL(h(X), Y)] \geq \E_{P_{{\rm s}}}[\cL(h_{{\rm s}}(X), Y)]
	\end{equation}
	for any $h(\cdot)$ with $\cI_{g} \subset \cT_{h}$ and for any $h_{{\rm s}}(\cdot) \in \cH_{{\rm s}}$. If $\cI_{g} \subset \cT_{h}$, according to Lemma \ref{lem: essential invariant set}, there exists $v(\cdot)$ such that $h(\cdot) = (v \circ g)(\cdot)$. By the definition of $h_{{\rm s}}(\cdot)$, there also exists $\phi_{{\rm s}}(\cdot)$ satisfying $h_{{\rm s}}(\cdot) = (\phi_{{\rm s}}\circ g)(\cdot)$ and $\phi_{{\rm s}}(w) \in \mathop{\arg\min}_{z} \E_{P_{{\rm s}}}[\cL(z, Y)\mid g(X) = w]$ for almost every $w$. Thus, we have
	\begin{equation}
	\begin{aligned}
	\E_{P_{{\rm s}}}[\cL(h(X), Y)\mid g(X) = w] & = \E_{P_{{\rm s}}}[\cL(v(w), Y)\mid g(X) = w]\\
	& \geq \E_{P_{{\rm s}}}[\cL(\phi_{{\rm s}}(w), Y)\mid g(X) = w] \\
	&  \geq \E_{P_{{\rm s}}}[\cL(h_{{\rm s}}(X), Y)\mid g(X) = w] \quad a.s.
	\end{aligned}\nonumber
	\end{equation}
	Then (\ref{eq: proof-constrained opt}) follows by taking expectation.
	
	Next we show the opposite inclusion to prove (\ref{eq: invariant feature problem-copy}). Suppose $h_{*}(\cdot)$ is a solution to the optimization problem in (\ref{eq: invariant feature problem-copy}). Then according Lemma~\ref{lem: essential invariant set}, there is some $v_{*}(\cdot)$ such that $h_{*}(\cdot) = (v_{*}\circ g)(\cdot)$. Let $h_{{\rm s}}(\cdot) = (\phi_{{\rm s}}\circ g)(\cdot) \in \cH_{{\rm s}}$. Then 
	\begin{equation}\label{eq: lower bound}
	\begin{aligned}
	\E_{P_{{\rm s}}}[\cL(h_{*}(X), Y)\mid g(X) = w]
	&= \E_{P_{{\rm s}}}[\cL(v_{*}(w), Y)\mid g(X) = w] \\
	& \geq  \E_{P_{{\rm s}}}[\cL(\phi_{{\rm s}}(w), Y)\mid g(X) = w] \\
	& =  \E_{P_{{\rm s}}}[\cL(h_{{\rm s}}(X), Y)\mid g(X) = w]  \quad a.s.,
	\end{aligned}
	\end{equation}
	by definition. 
	Because $h_{*}(\cdot)$ is a solution to the minimization problem, we have
	\begin{equation}
	\E_{P_{{\rm s}}}[\cL(h_{*}(X), Y)] = \E_{P_{{\rm s}}}[\cL(h_{{\rm s}}(X), Y)]. \nonumber
	\end{equation} 
	Combining this with (\ref{eq: lower bound}), we have
	\begin{equation}
	\E_{P_{{\rm s}}}[\cL(h_{*}(X), Y)\mid g(X) = w] \leq \E_{P_{{\rm s}}}[\cL(h_{{\rm s}}(X), Y)\mid g(X) = w] \quad a.s.
	\end{equation} 
	This implies 
	\begin{equation}
	\begin{aligned}
	\E_{P_{{\rm s}}}[\cL(v_{*}(w), Y)\mid g(X) = w]
	& \leq \E_{P_{{\rm s}}}[\cL(\phi_{{\rm s}}(w), Y)\mid g(X) = w]\\
	& = \min_{z}  \E_{P_{{\rm s}}}[\cL(z, Y)\mid g(X) = w]
	\quad a.s.\nonumber
	\end{aligned}
	\end{equation} 
	Thus, we conclude that $v_{*}(w) \in \mathop{\arg\min}_{z}  \E_{P_{{\rm s}}}[\cL(z, Y)\mid g(X) = w]$.
\end{proof}
\section{More Experimental Results}
\subsection{Toy Example and Simulation}\label{app: toy example}
In the following toy example, we are able to construct an explicit formulation of the causal essential invariant set. 
\begin{example}\label{eg: toy} \emph{
		Let $X$ be a non-singular $2\times 2$ matrix and $X^{(j)}$ be the $j$-th column of $X$ for $j=1,2$. Suppose that $g(X)$ is the area of the triangle formed by the two points $X^{(1)}$, $X^{(2)}$ and the origin. Then it is not hard to show that $\{T_{R,\theta}(\cdot), T_{S,a}(\cdot),T_{M}(\cdot), T_{P}(\cdot), T_{I}(\cdot)\mid  \theta \in \ [0,\pi/4], \ a\in [2/3,3/2]\}$ is  an essential invariant set of $g(\cdot)$, where 
		\begin{equation}
		\begin{array}{ll}
		T_{R,\theta}(X) = \left(
		\begin{array}{cc}
		\cos \theta & -\sin\theta\\
		\sin\theta  & \cos\theta 
		\end{array}
		\right)X , 
		&T_{S,a}(X) = X\left(
		\begin{array}{cc}
		a & 0\\
		0  & a^{-1} 
		\end{array}
		\right),\\ 
		T_M(X) = X\left(
		\begin{array}{cc}
		-1 & 0\\
		0  & 1 
		\end{array}
		\right), 
		&T_{P}(X) = X\left(
		\begin{array}{cc}
		1 & 1\\
		0  & 1 
		\end{array}
		\right), \\ 
		T_{I}(X) = X\left(
		\begin{array}{cc}
		-1 & 0\\
		0  & -1 
		\end{array}
		\right). &
		\end{array} \nonumber
		\end{equation}
		Here $T_{R,\theta}(\cdot)$ rotates the triangle with $\theta$ degree clockwise, and $T_{S,a}(\cdot)$ scales the two edges (one connects $X^{(1)}$ to the origin and the other connects $X^{(2)}$ to the origin)  of the triangle with $a$ and $a^{-1}$ times, respectively. $T_{M}(\cdot)$ mirrors the triangle with respect to the x-axis. $T_{P}(\cdot)$ transforms the triangle to another triangle with same base and height, and $T_{I}(\cdot)$ transforms the the triangle to another one that  is symmetric  with respect to the origin. All these transformations are known to keep the triangle area unchanged based on  elementary geometry.}
\end{example}
\par
Now we verify the effectiveness of the proposed method in the main body  using this example. 
\paragraph{Data.} We consider the following data generation process:
\begin{equation} \label{eq: DGP}
\begin{aligned}
&X^{(1)}\sim N(0, I_{2}), \ X^{(2)}\sim N(0, 2I_{2}),\ X = (X^{(1)}, X^{(2)}),\\
& \epsilon \sim N(0,1), \ \eta = \frac{a\Phi^{-1}(\pi^{-1}\alpha) + \epsilon}{\sqrt{a^{2} + 1}},\\
& Y = |\det(X)| + \eta, 
\end{aligned}
\end{equation}
where $I_{2}$ is the identity matrix of order $2$, $\Phi(\cdot)$ is the cumulative distribution function of standard normal distribution. 
In this data generation process, $|\det(X)|$ is the area of the triangle formed by $X^{(1)}$, $X^{(2)}$ and the origin, and is the causal feature in this example. Here $\alpha$ is the angle between $(X^{(1)} + X^{(2)})/2$ and x-axis, and is  correlated with $Y$ in certain domains, with $a$ a parameter that reflects this correlation. However, this correlation is a spurious correlation that changes across domains, i.e., $a$ is set to be different in different domains.
In the training population, we pick $a=-3$. We then generate i.i.d.~samples of size $1,000$, denoted by $\{(Y_{i}, X_{i})\}_{i=1}^{1000}$, and  train a model $h(X,\beta)$ with parameter $\beta$ to predict $Y$ based on these generated samples.
\paragraph{Model.} For any $2\times 2$ matrix
\begin{equation}
X = \left(\begin{array}{cc}
X_{11}&X_{12}\\
X_{21}&X_{22}
\end{array}\right), \nonumber
\end{equation}
let 
\begin{equation}
\begin{aligned}
v(X)& = (1, X_{11}, X_{21}, X_{12}, X_{22}, X_{11}^{2}, X_{21}^{2}, X_{12}^{2}, X_{22}^{2},\\
&\quad~~~X_{11}X_{21}, X_{11}X_{12}, X_{11}X_{22}, X_{21}X_{12}, X_{21}X_{22}, X_{12}X_{22})^{\T}.\nonumber
\end{aligned}
\end{equation}
The model is
\begin{equation}
h_{\beta}(X) = {\rm ReLU}(\beta_{[1]}^{\T}v(X)) + \beta_{[2]}^{\T}v(X), \nonumber
\end{equation}
where $\beta = (\beta_{[1]}^{\T}, \beta_{[2]}^{\T})^{\T}$ is the model parameter. We pick this model because we have known that $|\det(X)|$ is a function of $v(X)$, and  there is some $\beta^{*}$ such that $h_{\beta^{*}}(X) = |\det(X)|$.

\paragraph{Method.} Based on the essential invariant set given in Example \ref{eg: toy}, we define five invariant transformations
\begin{equation}
\begin{array}{ll}
T_{1}(X) = \left(
\begin{array}{cc}
\cos\frac{\pi}{12} & -\sin\frac{\pi}{12}\\
\sin\frac{\pi}{12}  & \cos\frac{\pi}{12}
\end{array}
\right)X , 
&T_{2}(X) = X\left(
\begin{array}{cc}
1.1 & 0\\
0  & 1.1^{-1} 
\end{array}
\right),\\ 
T_3(X) = X\left(
\begin{array}{cc}
-1 & 0\\
0  & 1 
\end{array}
\right), 
&T_{4}(X) = X\left(
\begin{array}{cc}
1 & 1\\
0  & 1 
\end{array}
\right), \\ 
T_{5}(X) = X\left(
\begin{array}{cc}
-1 & 0\\
0  & -1 
\end{array}
\right). &
\end{array}\nonumber
\end{equation}
For ease of notation, we let $T_{0}(X) = X$ be the identity transformation.
We learn the model parameter by minimizing  four different loss functions, namely, the empirical risk
\begin{equation}
\frac{1}{n}\sum_{i=1}^{n}(Y_{i} - h_{\beta}(X_{i}))^{2}, \nonumber
\end{equation}
the average risk over different transformations
\begin{equation}
\frac{1}{n}\sum_{k=0}^{5}\sum_{i=1}^{n}(Y_{i} - h_{\beta}(T_{k}(X_{i})))^{2},
\nonumber
\end{equation}
the maximal risk over different transformations
\begin{equation}
\max_{k=0,\dots,5}\left\{\frac{1}{n}\sum_{i=1}^{n}(Y_{i} - h_{\beta}(T_{k}(X_{i})))^{2}\right\},\nonumber
\end{equation}
and the RICE loss function
\begin{equation}
\frac{1}{n}\sum_{i=1}^{n}(Y_{i} - h_{\beta}(X_{i}))^{2}+\lambda \max_{k=0,\ldots,5}\left\{\frac{1}{n}\sum_{i=1}^{n}(h_{\beta}(X_{i}) - h_{\beta}(T_{k}(X_{i}))^{2}\right\},\nonumber
\end{equation}
where $n=1000$. For the given quantities $l_{0}, \cdots, l_{5}$, we replace the maximum  $\max_{k=0,\dots,5}\{l_{k}\}$  in the above losses with the softmax weighting quantity 
$\sum_{k=0}^{5}\exp(0.2l_{k})l_{k}/\sum_{k=0}^{5}\exp(0.2l_{k})$
for ease of computation in the implementation of RICE.
\paragraph{Results.} The resulting model is evaluated on i.i.d.~sample generated following the data generation process (\ref{eq: DGP}) with different $a$. The following figure plots the squared prediction error of the four methods on test data  with different values of $a$. Each reported value is the average over $200$ simulations.

\begin{figure}[h]
	\centering
	\includegraphics[scale=0.4]{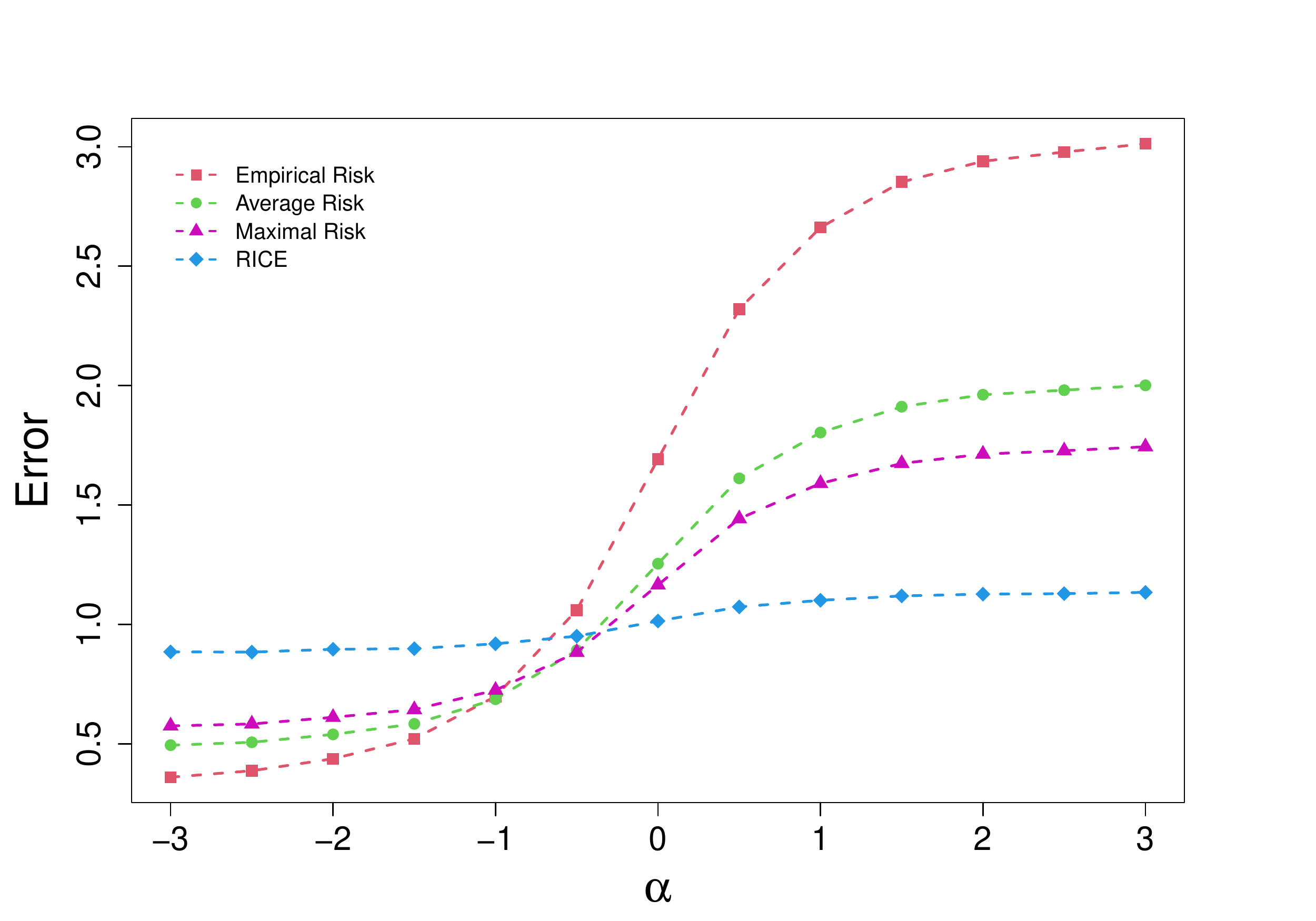}
	\caption{Squared prediction error on test data from distributions with different values of $a$.}
\end{figure}

It can be seen that when the test distribution has similar spurious correlations as the training population, minimizing the empirical risk performs the best among the four methods. However, it performs the worst if an opposite spurious correlation appears in the test population. The RICE algorithm has the best worst-case performance, which is consistent with our theoretical analysis. Moreover, the RICE algorithm seems successfully capture the invariant causal mechanism across different environments, as its prediction errors under different test distributions are stable and close to the variance of the intrinsic error $\eta$. 
\subsection{Hyperparameters}\label{app: hyperparameters}
We summarize the hyperparameters of the proposed RICE for \texttt{C-MNIST}, \texttt{PACS}, and \texttt{VLCS} datasets in Table \ref{tbl:hyper}. The learning rate is decayed by $0.2$ at epoch $6, 12$, and $20$. 
\begin{table*}[t!]
	\caption{Hyperparameters of the proposed RICE on \texttt{C-MNIST}, \texttt{PACS}, and \texttt{VLCS}.}
	\label{tbl:hyper}
	\centering
	{
		{
			\begin{tabular}{c c c c}
				\hline
				Dataset & \texttt{C-MNIST}    & \texttt{PACS}   &   \texttt{VLCS} \\
				\hline
				Learning Rate &  0.1      & 5e-5  &  5e-5 \\
				Batch Size    &  128      & 32   &   32  \\
				Weight Decay  &  5e-4     & 0    &   0   \\
				Drop Out      &  0        & 0.1  &   0.1 \\
				Epoch         &  20       & 20   &   20 \\
				$\lambda_{0}$ &  0.25     & 0.5 &    0.5  \\
				$\beta_{1}$   &  0.9      & 0.9  &   0.9  \\
				$\beta_{2}$   &  0.999    & 0.999&   0.999 \\ 
				\hline
	\end{tabular}}}
\end{table*}
\subsection{Ablation Study}\label{app: ablation}
In Section \ref{sec:experiment}, for the experiments on \texttt{PACS} and \texttt{VLCS}, we collect training data from several domains for the proposed RICE. However, our theory in Section \ref{subsec: OOD&CIT}  requires only a single domain. Thus, in this subsection, we study the performance of RICE with single domain training data. 
\par
Our experiments are conducted on both \texttt{PACS} and \texttt{VLCS}. All the hyperparameters are set to be same with those in Section \ref{sec:experiment}, except the number of training domains---we only use single domain data  and hence less training samples for each single experiment. For example, for \texttt{PACS}, if the test domain is sketch, then we  run  RICE on training data from one of the three other domains (photo, art and cartoon) and report the accuracy on the test domain. To run RICE, the data  generated by CycleGAN  are used as augmented data and in the regularization term.  For a fair comparison, we do not use the CycleGAN that transfer from training domain to test domain and adopt  similar experimental settings for ERM.
\par
\begin{figure*}[t!]\centering
	\subcaptionbox{Photo}{
		\includegraphics[width=0.43\textwidth]{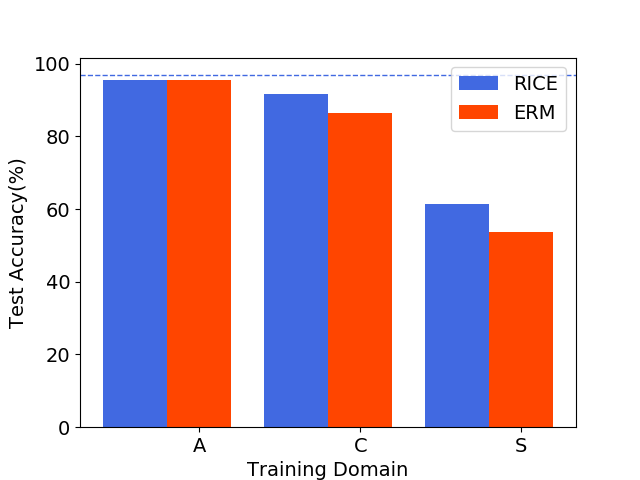}}
	\subcaptionbox{Art}{
		\includegraphics[width=0.43\textwidth]{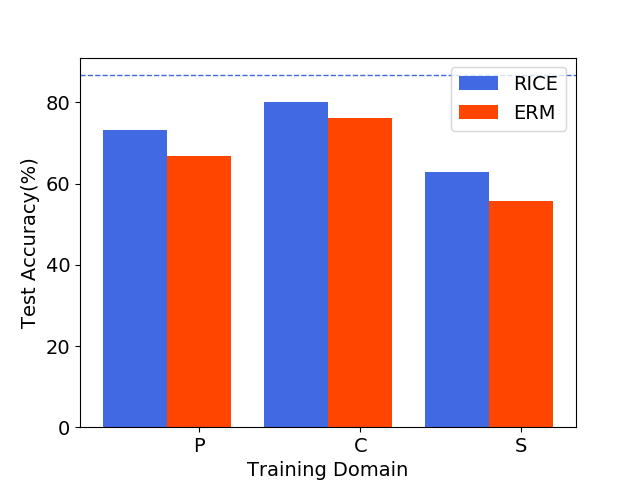}} \\
	\vspace{-1em}
	\subcaptionbox{Cartoon}{
		\includegraphics[width=0.43\textwidth]{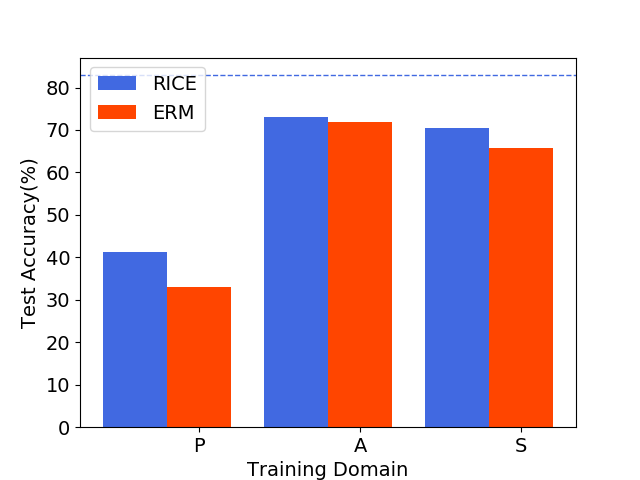}}
	\subcaptionbox{Sketch}{
		\includegraphics[width=0.43\textwidth]{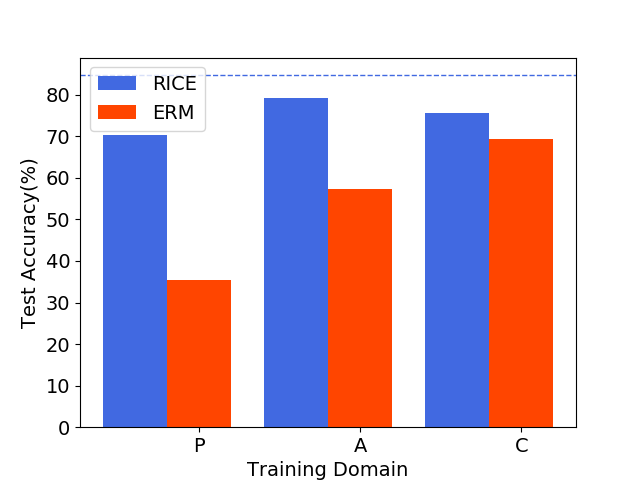}}\\
	\vspace{-1em}
	\subcaptionbox{VOC2007}{
		\includegraphics[width=0.43\textwidth]{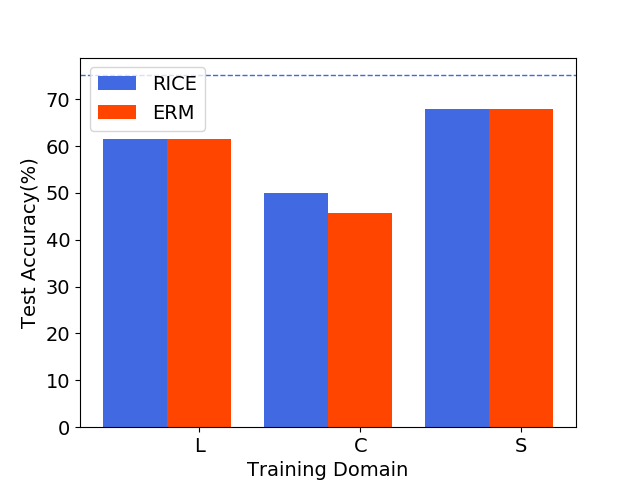}}
	\subcaptionbox{LabelMe}{
		\includegraphics[width=0.43\textwidth]{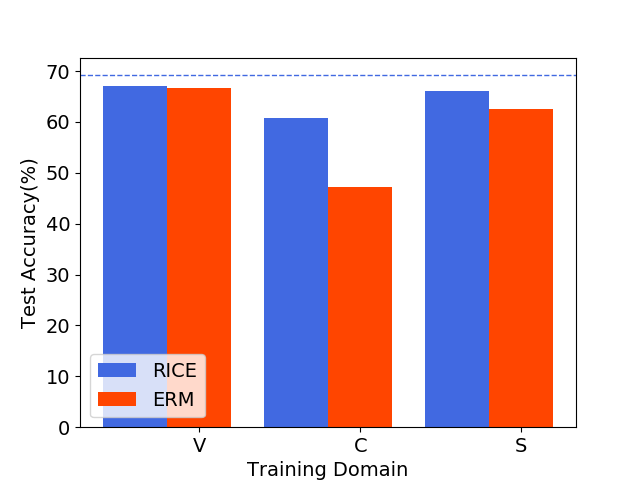}}
	\\
	\vspace{-1em}
	\subcaptionbox{Caltech101}{
		\includegraphics[width=0.43\textwidth]{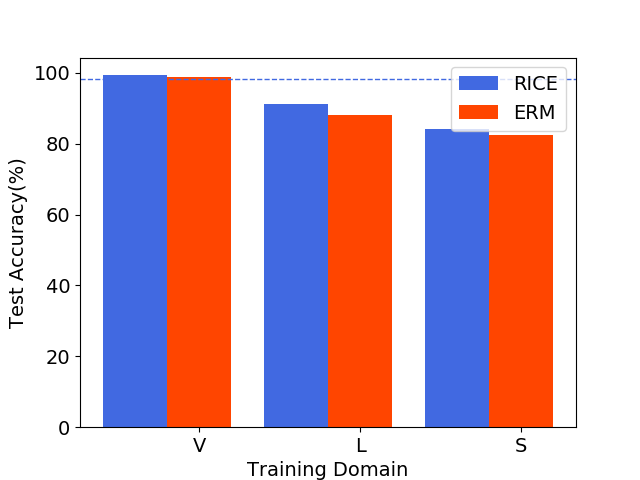}}
	\subcaptionbox{SUN09}{
		\includegraphics[width=0.43\textwidth]{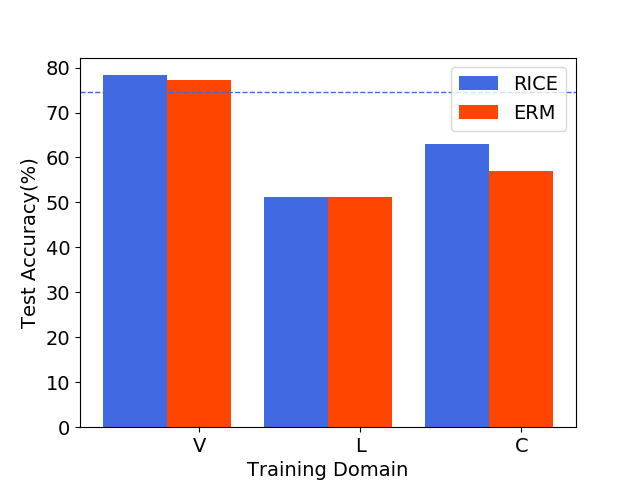}}
	\caption{Performance of RICE and ERM on the \texttt{PACS} (a-d) and \texttt{VLCS} (e-h) datasets with training data from single domains. Figure title indicates the  test domain, and the blue dashed line represents the test accuracy when the training data are from three domains, as reported in Section~\ref{sec:experiment}.}
	\label{fig: ablation1}
\end{figure*}

The results are summarized in Figure \ref{fig: ablation1}. We can see that RICE performs much better than the baseline method ERM, which verifies our theoretical conclusions in Theorem \ref{thm: alternative problem 2}. Besides, the test accuracy on the target domain can be quite high even when the model is trained using data from a single domain. For example, on \texttt{VLCS} dataset, when test data is from  SUN09 domain, the model trained on VOC2007 domain even exhibits a better OOD generalization than the model trained on data from three domains. This implies that, for OOD generalization problem, the number of domains may not be crucial to the performance as long as some representative CITs are available.

\subsection{Generated Data}\label{app: generated data}
\begin{figure*}[t!]\centering
	\subcaptionbox{Training data in \texttt{C-MNIST}}{
		\includegraphics[width=0.35\textwidth]{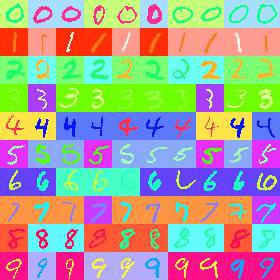}}
	\hspace{0.2in}
	\subcaptionbox{Test data in \texttt{C-MNIST}}{
		\includegraphics[width=0.35\textwidth]{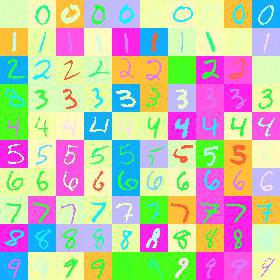}}
	\caption{Images of the \texttt{C-MNIST} dataset.}
	\label{fig: c-mnist}
\end{figure*}

\begin{figure*}[t!]\centering
	\vspace{-0.1in}
	\subcaptionbox{original domain: photo}{
		\includegraphics[width=0.35\textwidth]{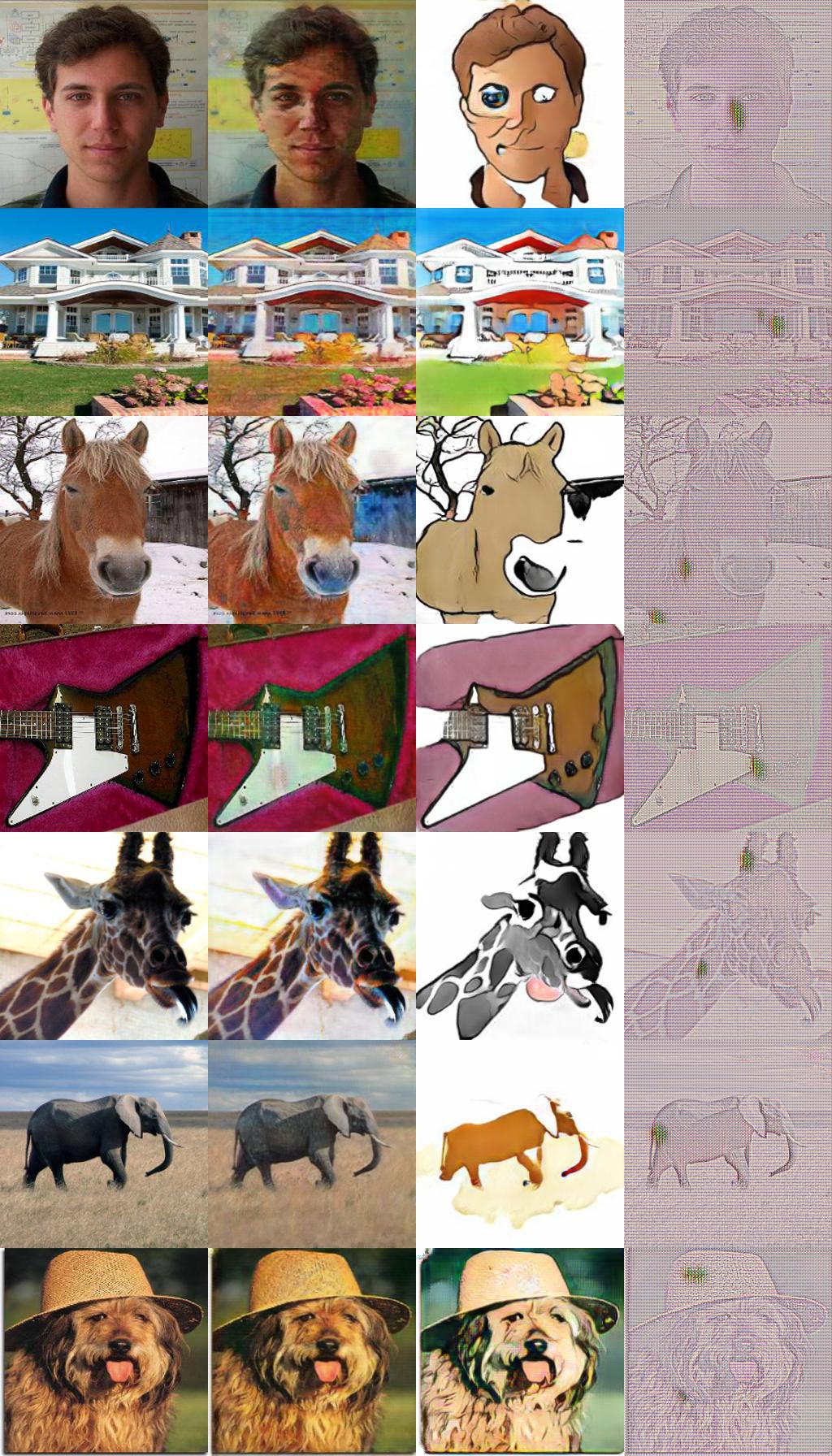}}
	\hspace{0.2in}
	\subcaptionbox{original domain: art}{
		\includegraphics[width=0.35\textwidth]{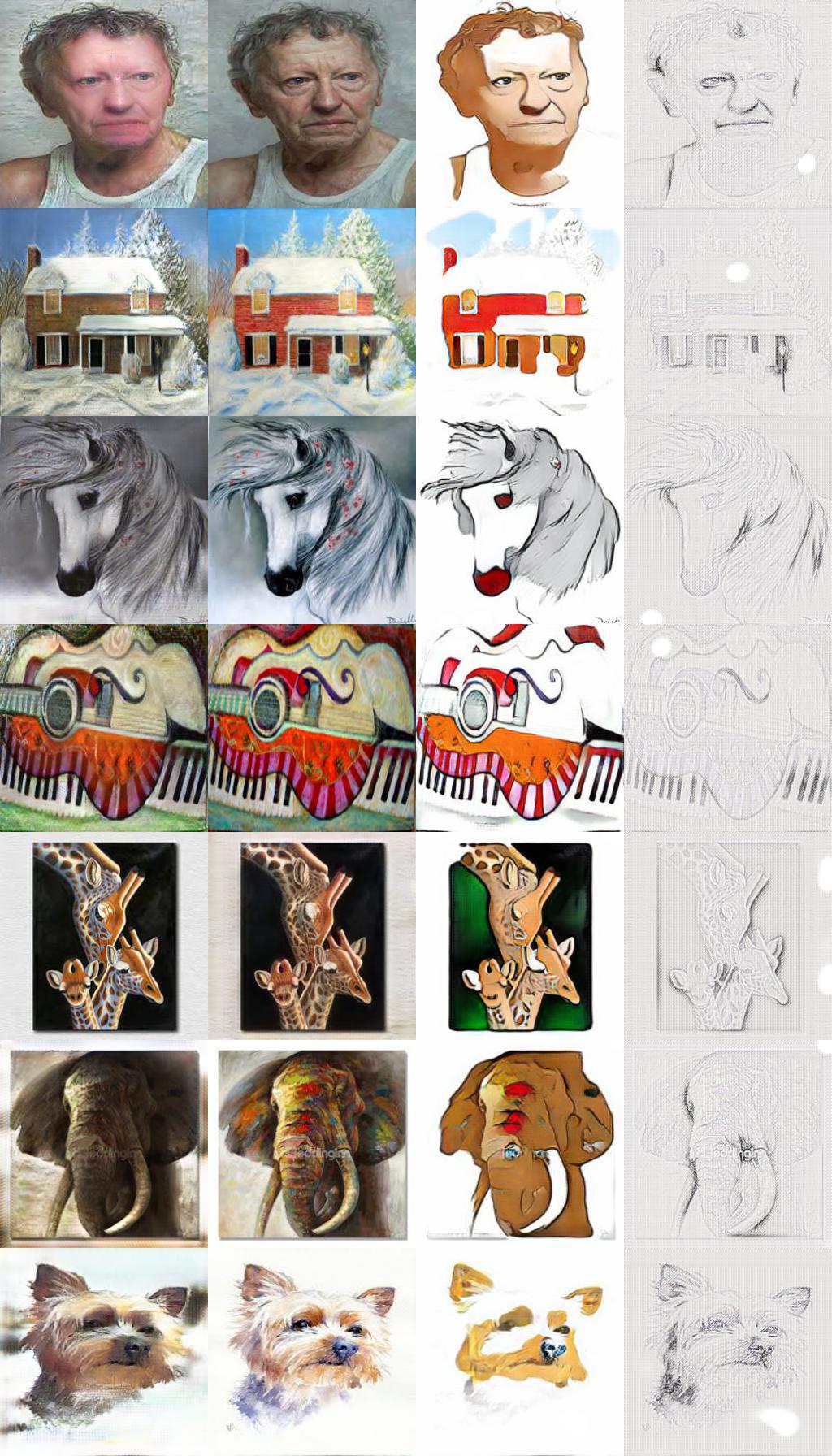}}
	\\
	\subcaptionbox{original domain: cartoon\label{fig: pacs-c}}{
		\includegraphics[width=0.35\textwidth]{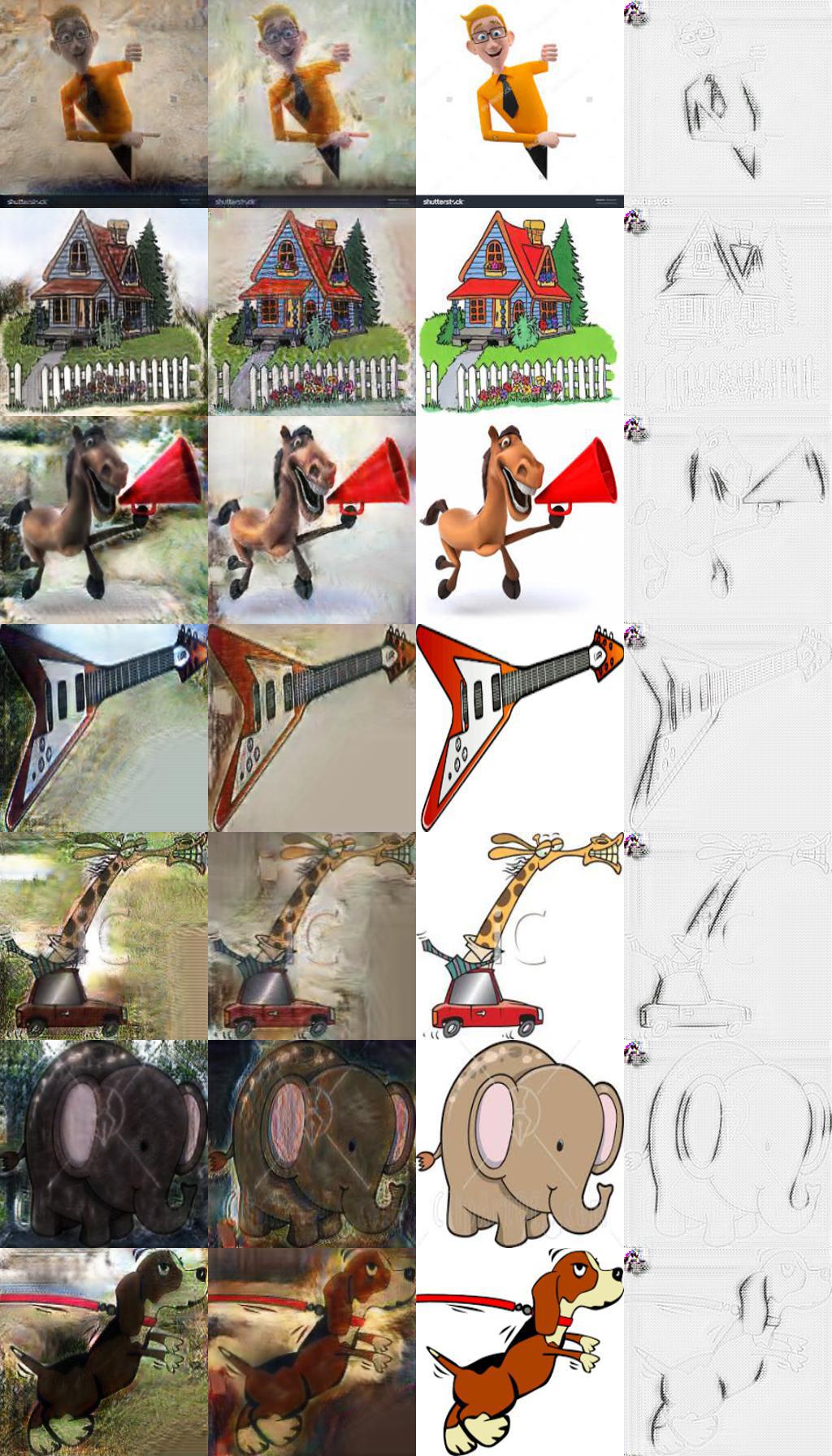}}
	\hspace{0.2in}
	\subcaptionbox{original domain: sketch}{
		\includegraphics[width=0.35\textwidth]{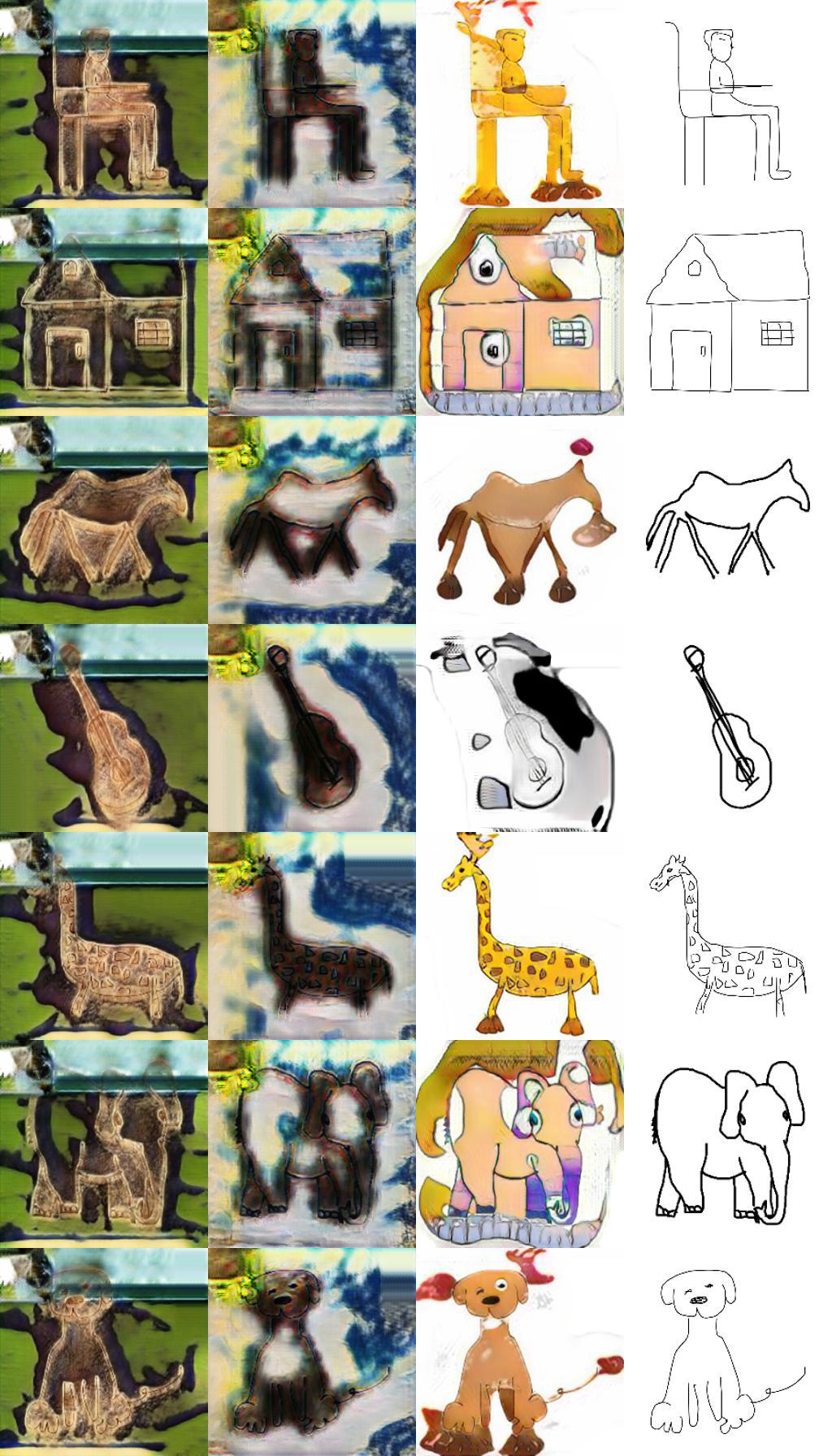}}
	\caption{Synthetic data of \texttt{PACS} generated by CycleGAN. Columns from  left to right  correspond to domains of \{photo, art, cartoon, sketch\}, respectively. Figure title indicates the domain of original data, based on which the data of the rest domains in the figure are generated by CycleGAN.}
	\label{fig: pacs}
\end{figure*}

Our experiments in the main body involve generating causally invariant images. In this subsection, we present visualizations of some generated images for a better understanding of the proposed algorithm.
\paragraph{\texttt{C-MNIST}}  Figure \ref{fig: c-mnist} shows some \texttt{C-MNIST} images. As seen from the training set, there exist spurious correlations between the colors of the foreground or background and the category. However, the correlation disappears in the test set, as the foreground and background colors are randomly assigned. 
\paragraph{\texttt{PACS}} We also present some transformed data from \texttt{PACS} dataset generated by CycleGAN. The CycleGAN is used to simulate CITs as we have clarified the main body of this paper. As the data in \texttt{PACS} come from $7$ categories, for each category we pick $4$ pictures respectively from domains \{photo, art, cartoon, sketch\}.  The transformed images are shown in Figure \ref{fig: pacs}, where the columns correspond to the styles of \{photo, art, cartoon, sketch\}, respectively.   
\par
Let us look at these generated data over different domains. For the generated images of the photo domain (the first column), the trained CycleGAN tends to alter its color of foreground and add a background, especially when the original images are from the cartoon and sketch domains. Similar trends exhibit in the generated data of the art domain (the second column). In contrast to the two aforementioned domains, the generated cartoon data in the third column remove the background (if exists) while keep or alter the color of the foreground. The generated sketch data (the fourth column) are more likely to be a grayscale view of the original images. However, for each generated image, the shape of its foreground (i.e., the casual feature to decide the category) does not change when we vary the domains.
\par
\begin{figure*}[t!]\centering
	\subcaptionbox{original domain: VOC2007}{
		\includegraphics[width=0.35\textwidth]{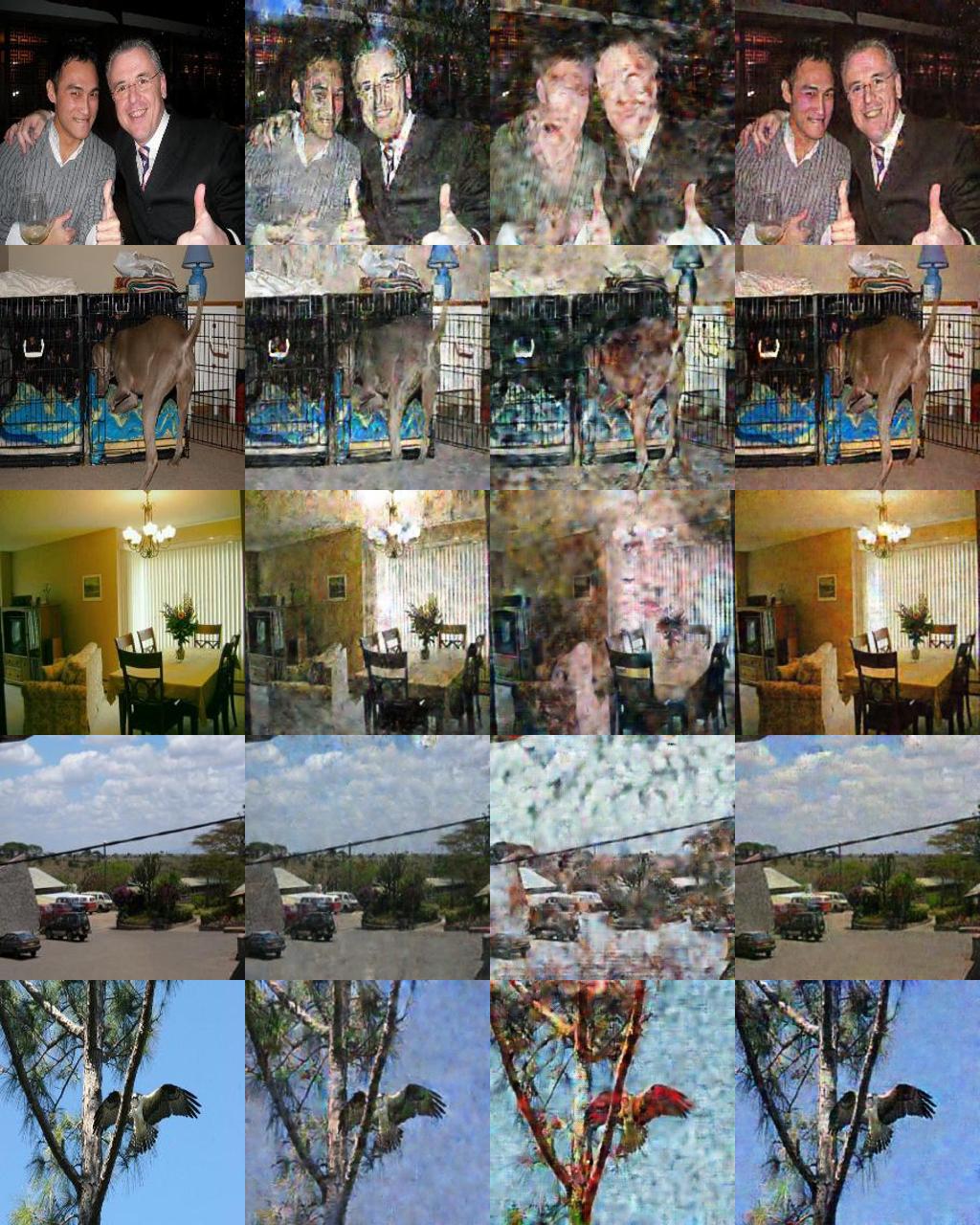}}
	\hspace{0.2in}
	\subcaptionbox{original domain: LabelMe}{
		\includegraphics[width=0.35\textwidth]{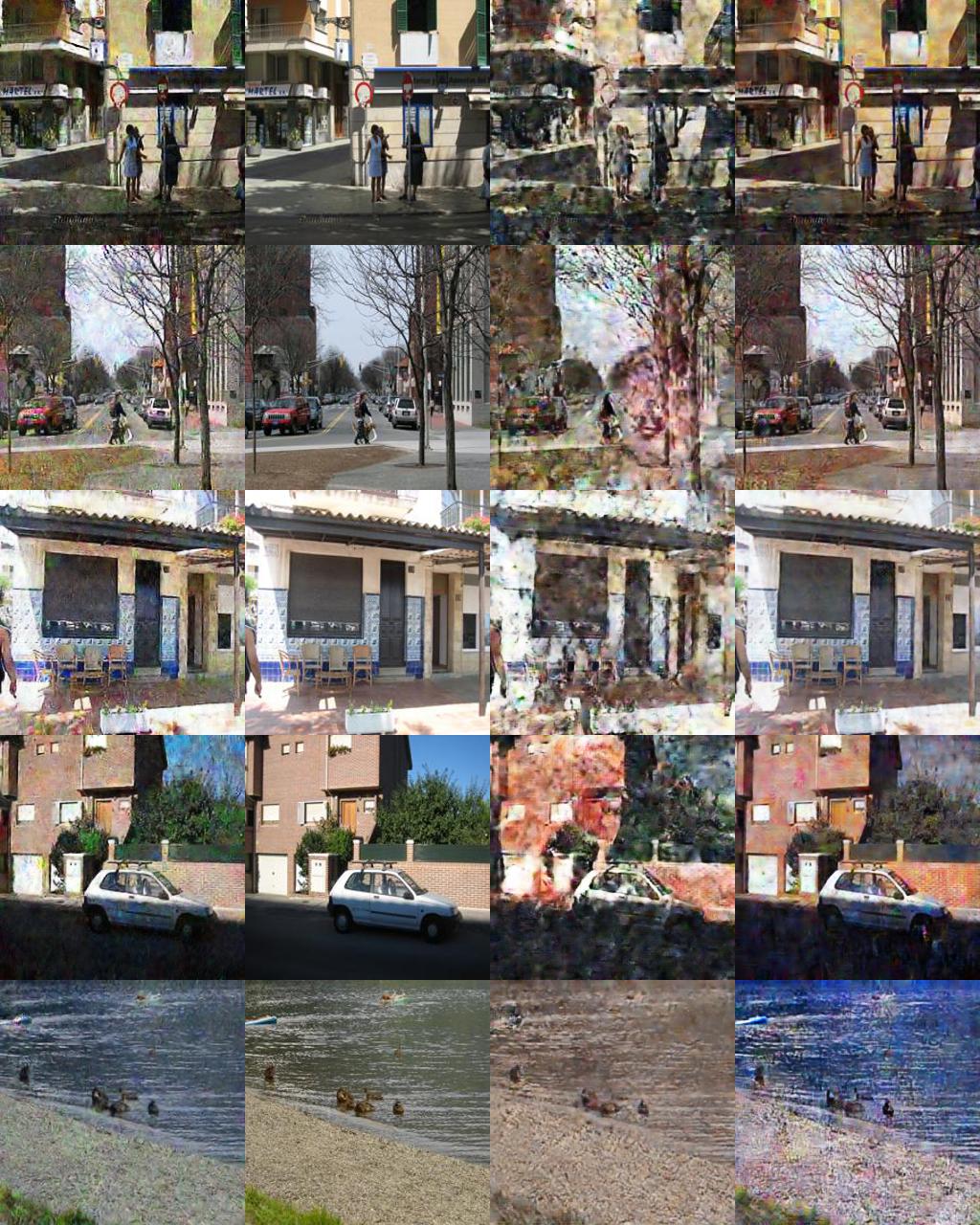}}
	\\
	\subcaptionbox{original domain: Caltech101}{
		\includegraphics[width=0.35\textwidth]{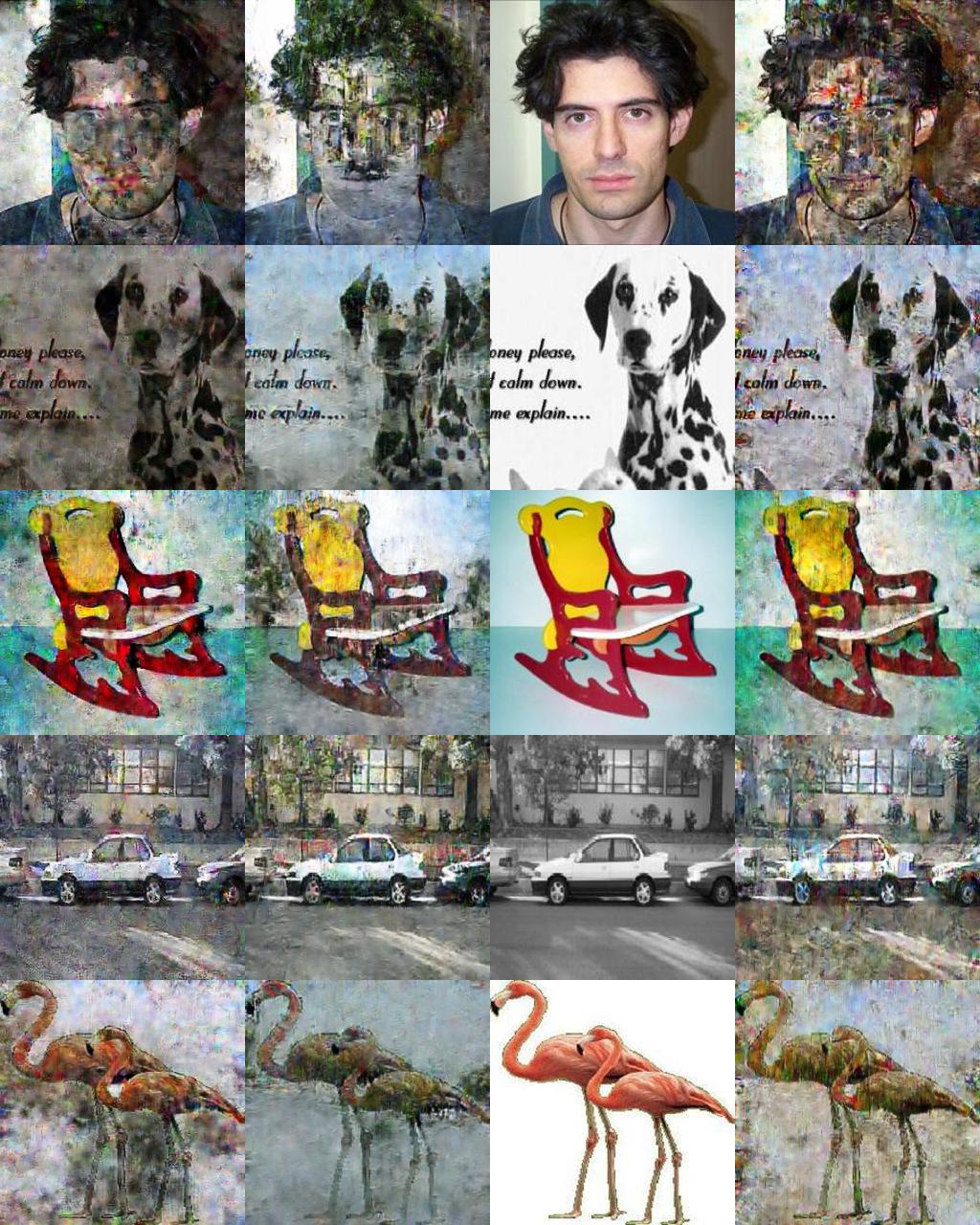}}
	\hspace{0.2in}
	\subcaptionbox{original domain: SUN09 }{
		\includegraphics[width=0.35\textwidth]{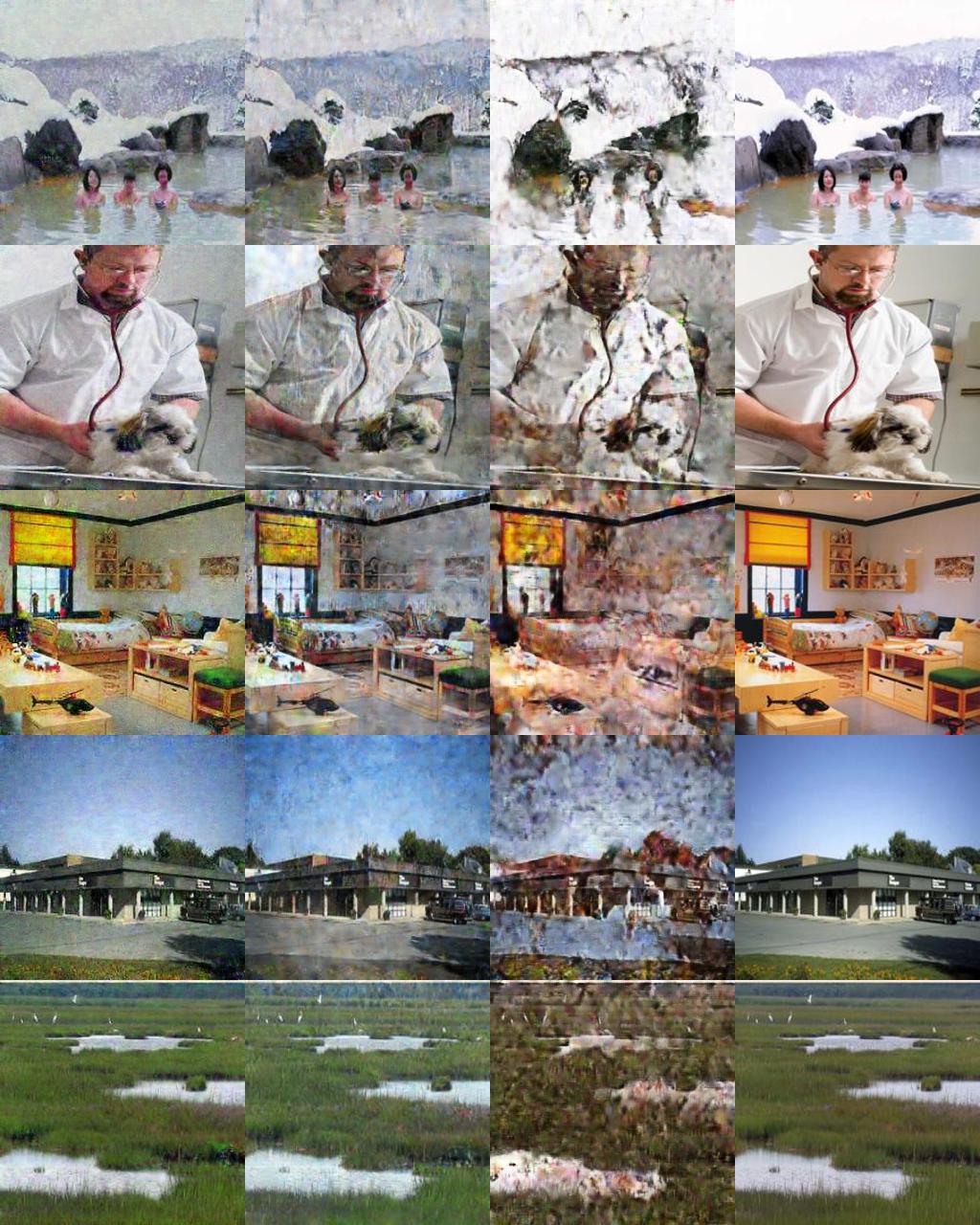}}
	\caption{Synthetic data of \texttt{VLCS} generated by CycleGAN. Columns from  left to right  correspond to domains of \{VOC2007, LabelMe, Caltech101, SUN09\}, respectively. Figure title indicates the domain of original data, based on which the data of the rest domains in the figure are generated by CycleGAN.}
	\label{fig: vlcs}
\end{figure*}

The proposed algorithm RICE  regularizes the model to encourage the model to be invariant under the CITs, i.e., invariant to the changes of spurious features. This enables the model to be robust to the misleading signal from spurious features and to make predictions via the casual feature. For example, for the dog images in the last row of Figure \ref{fig: pacs-c}, which are generated from the images of cartoon style (the third column), the generated dog image of photo style (the first column)  has a grass background. However,  RICE requires the model to exhibit similar outputs for the two images, hence breaking the spurious correlation between dog and grass. 
\paragraph{\texttt{VLCS}} Similar to \texttt{PACS}, we present some of the domain transformed data from \texttt{VLCS} dataset generated by CycleGAN. We pick $4$ pictures respectively from domains \{VOC2007, LabelMe, Caltech101, SUN09\} for each of the $5$ categories in \texttt{VLCS}. Then we vary the domains of these picked data using the trained CycleGAN models. The transformed data are visualized in Figure \ref{fig: vlcs}. 
\par
The generated \texttt{VLCS} images exhibit similar behaviors as \texttt{PACS}. Specifically, for a given image from a certain domain, the CycleGAN model tends to deterministically vary the color of the background according to the domains. Thus, the reasoning about the effectiveness of RICE on \texttt{PACS} also applies here.  

\subsection{Benchmark algorithms}\label{app: benchmark algo}
\begin{itemize}
	\item Empirical Risk minimization (ERM) pools together the data from all the domains  and then minimizes the empirical loss to train the model. Notice that here an \texttt{ImageNET} pre-trained model is used.
	\item Marginal Transfer Learning \cite{blanchard2021domain} use the mean embedding of the feature distribution in each domain as an input of the classifier.
	\item Group Distributionally Robust Optimization (GroupDRO) \cite{sagawa2020distributionally} minimizes the largest loss across different domains.
	\item Domain-Adversarial Neural Networks (DANN) \cite{ganin2016domain} use adversarial networks to match the feature distribution in different domains.
	\item Invariant Risk Minimization (IRM )\cite{arjovsky2019invariant} learns a feature representation such that the optimal classifiers on top of the representation is the same across the domains.
\end{itemize}